\documentclass{article}

\usepackage{microtype}
\usepackage{graphicx}
\usepackage{booktabs} 

\usepackage{hyperref}

\usepackage[accepted]{icml2023}

\usepackage{amsmath}
\usepackage{amssymb}
\usepackage{mathtools}
\usepackage{amsthm}

\usepackage[capitalize,noabbrev]{cleveref}

\theoremstyle{plain}
\newtheorem{theorem}{Theorem}[section]

\theoremstyle{definition}
\newtheorem{definition}[theorem]{Definition}

\theoremstyle{remark}

\usepackage{dsfont}
\usepackage{bbding}
\usepackage{adjustbox}
\usepackage[para,online,flushleft]{threeparttable}
\usepackage{multicol}
\usepackage{multirow}
\usepackage{caption}
\usepackage{subcaption}

\usepackage[textsize=tiny]{todonotes}

\usepackage{url}
\usepackage{breakurl}
\usepackage[breaklinks]{}
\usepackage{xcolor}

\icmltitlerunning{Are Diffusion Models Vulnerable to Membership Inference Attacks?}

\usepackage{amsmath,amsfonts,bm, bbm}
\usepackage{xparse}

\DeclareDocumentCommand\W{ g g }{%
        \IfNoValueTF {#1} {\mathbf{W}} {
            \IfNoValueTF {#2} {\mathbf{W}^{(#1)}}{\mathbf{W}^{(#1)}_{#2}}
        }
}

\DeclareDocumentCommand\bias{ g g }{%
        \IfNoValueTF {#1} {\mathbf{b}} {
            \IfNoValueTF {#2} {\mathbf{b}^{(#1)}}{\mathbf{b}^{(#1)}_{#2}}
        }
}

\DeclareDocumentCommand\betavar{ g g }{%
        \IfNoValueTF {#1} {\bm{\beta}} {
            \IfNoValueTF {#2} {{\bm{\beta}^{(#1)}}{}}{\bm{\beta}^{(#1)}_{#2}}
        }
}

\DeclareDocumentCommand\xivar{ g g }{%
        \IfNoValueTF {#1} {\bm{\xi}} {
            \IfNoValueTF {#2} {{\bm{\xi}^{(#1)}}{}}{\bm{\xi}^{(#1)}_{#2}}
        }
}

\DeclareDocumentCommand\xivarn{ g g }{%
        \IfNoValueTF {#1} {\bm{\xi^-}} {
            \IfNoValueTF {#2} {\bm{\xi^-}^{+(#1)}}{\bm{\xi^-}^{+(#1)}_{#2}}
        }
}

\DeclareDocumentCommand\xivarp{ g g }{%
        \IfNoValueTF {#1} {\bm{\xi^+}} {
            \IfNoValueTF {#2} {\bm{\xi^+}^{+(#1)}}{\bm{\xi^+}^{+(#1)}_{#2}}
        }
}

\DeclareDocumentCommand\nuvar{ g g }{%
        \IfNoValueTF {#1} {\bm{\nu}} {
            \IfNoValueTF {#2} {{\bm{\nu}^{(#1)}}{}}{\bm{\nu}^{(#1)}_{#2}{}}
        }
}

\DeclareDocumentCommand\hnuvar{ g g }{%
        \IfNoValueTF {#1} {\bm{\hat{\nu}}} {
            \IfNoValueTF {#2} {{\bm{\hat{\nu}}^{(#1)}}{}}{\bm{\hat{\nu}}^{(#1)}_{#2}{}}
        }
}

\DeclareDocumentCommand\muvar{ g g }{%
        \IfNoValueTF {#1} {\bm{\mu}} {
            \IfNoValueTF {#2} {{\bm{\mu}^{(#1)}}{}}{\bm{\mu}^{(#1)}_{#2}}
        }
}

\DeclareDocumentCommand\gammavar{ g g }{%
        \IfNoValueTF {#1} {\bm{\gamma}} {
            \IfNoValueTF {#2} {{\bm{\gamma}^{(#1)}}{}}{\bm{\gamma}^{(#1)}_{#2}}
        }
}

\DeclareDocumentCommand\lambdavar{ g g }{%
        \IfNoValueTF {#1} {\bm{\lambda}} {
            \IfNoValueTF {#2} {{\bm{\lambda}^{(#1)}}{}}{\bm{\lambda}^{(#1)}_{#2}}
        }
}

\DeclareDocumentCommand\tbetavar{ g g }{%
        \IfNoValueTF {#1} {{\bm{\tilde{\beta}}}} {
            \IfNoValueTF {#2} {{{\bm{\tilde{\beta}}}^{(#1)}}{}}{{{\bm{\tilde{\beta}}}^{(#1)}_{#2}}}
        }
}

\DeclareDocumentCommand\alphavar{ g g }{%
        \IfNoValueTF {#1} {\bm{\alpha}} {
            \IfNoValueTF {#2} {{\bm{\alpha}^{(#1)}}}{\bm{\alpha}^{(#1)}_{#2}}
        }
}

\DeclareDocumentCommand\D{ g g }{%
        \IfNoValueTF {#1} {\mathbf{D}} {
            \IfNoValueTF {#2} {\mathbf{D}^{(#1)}}{\mathbf{D}^{(#1)}_{#2}}
        }
}

\DeclareDocumentCommand\A{ g g }{%
        \IfNoValueTF {#1} {\mathbf{A}} {
            \IfNoValueTF {#2} {\mathbf{A}^{(#1)}}{\mathbf{A}^{(#1)}_{#2}}
        }
}

\DeclareDocumentCommand\AA{ g g }{
        \IfNoValueTF {#1} {\mathbf{\Omega}} {
            \IfNoValueTF {#2} {\mathbf{\Omega}(#1, #1)}{\mathbf{\Omega}(#1, #2)}
        }
}

\DeclareDocumentCommand\S{ g g }{%
        \IfNoValueTF {#1} {\mathbf{S}} {
            \IfNoValueTF {#2} {\mathbf{S}^{(#1)}}{\mathbf{S}^{(#1)}_{#2}}
        }
}

\DeclareDocumentCommand\K{ g g }{%
        \IfNoValueTF {#1} {\mathbf{K}} {
            \IfNoValueTF {#2} {\mathbf{K}^{(#1)}}{\mathbf{K}^{(#1)}_{#2}}
        }
}

\DeclareDocumentCommand\B{ g g }{%
        \IfNoValueTF {#1} {\mathbf{B}} {
            \IfNoValueTF {#2} {\mathbf{B}^{(#1)}}{\mathbf{B}^{(#1)}_{#2}}
        }
}

\DeclareDocumentCommand\lowerb{ g g }{%
        \IfNoValueTF {#1} {{\mathbf{\underline{b}}}} {
            \IfNoValueTF {#2} {{\mathbf{\underline{b}}}^{(#1)}}{{\mathbf{\underline{b}}}^{(#1)}_{#2}}
        }
}

\DeclareDocumentCommand\z{ g g }{%
        \IfNoValueTF {#1} {z} {
            \IfNoValueTF {#2} {z^{(#1)}}{z^{(#1)}_{#2}}
        }
}

\DeclareDocumentCommand\s{ g g }{%
        \IfNoValueTF {#1} {s} {
            \IfNoValueTF {#2} {s^{(#1)}}{s^{(#1)}_{#2}}
        }
}

\DeclareDocumentCommand\dom{ g g }{%
        \IfNoValueTF {#1} {\mathcal{S}} {
            \IfNoValueTF {#2} {\mathcal{S}_{#1}}{\mathcal{S}^{#1}_{#2}}
        }
}

\DeclareDocumentCommand\domlb{ g g }{%
        \IfNoValueTF {#1} {\mathsf{LB}} {
            \IfNoValueTF {#2} {\mathsf{LB}(\mathcal{#1})}{\mathsf{LB}(\mathcal{#1}_{#2})}
        }
}

\DeclareDocumentCommand\domub{ g g }{%
        \IfNoValueTF {#1} {\mathsf{UB}} {
            \IfNoValueTF {#2} {\mathsf{UB}(\mathcal{#1})}{\mathsf{UB}(\mathcal{#1}_{#2})}
        }
}

\DeclareDocumentCommand\uns{ g g }{%
        \IfNoValueTF {#1} {\tilde{s}} {
            \IfNoValueTF {#2} {\tilde{s}_{#1}}{s^{(#1)}_{#2}}
        }
}

\DeclareDocumentCommand\ub{ g g }{%
        \IfNoValueTF {#1} {u} {
            \IfNoValueTF {#2} {u^{(#1)}}{u^{(#1)}_{#2}}
        }
}

\DeclareDocumentCommand\lb{ g g }{%
        \IfNoValueTF {#1} {l} {
            \IfNoValueTF {#2} {l^{(#1)}}{l^{(#1)}_{#2}}
        }
}

\DeclareDocumentCommand\hz{ g g }{%
        \IfNoValueTF {#1} {\hat{z}} {
            \IfNoValueTF {#2} {\hat{z}^{(#1)}}{\hat{z}^{(#1)}_{#2}}
        }
}

\DeclareDocumentCommand\bu{ g g }{%
        \IfNoValueTF {#1} {\mathbf{u}} {
            \IfNoValueTF {#2} {\mathbf{u}^{(#1)}}{\mathbf{u}^{(#1)}_{#2}}
        }
}

\DeclareDocumentCommand\bl{ g g }{%
        \IfNoValueTF {#1} {\mathbf{l}} {
            \IfNoValueTF {#2} {\mathbf{l}^{(#1)}}{\mathbf{l}^{(#1)}_{#2}}
        }
}

\DeclareDocumentCommand\aaa{ g }{%
        \IfNoValueTF {#1} {\bm{a}} {
            {\bm{a}^{({#1})}}
        }
}

\DeclareDocumentCommand\haaa{ g }{%
        \IfNoValueTF {#1} {\bm{\hat{a}}} {
            {\bm{\hat{a}}^{({#1})}}
        }
}

\DeclareDocumentCommand\bbb{ g g }{%
        \IfNoValueTF {#1} {\mathbf{P}} {
            \IfNoValueTF {#2} {{\mathbf{P}_{#1}}}{{\mathbf{P}_{#1}^{({#2})}}}
        }
}

\DeclareDocumentCommand\hbbb{ g g }{%
        \IfNoValueTF {#1} {\mathbf{\hat{P}}} {
            \IfNoValueTF {#2} {{\mathbf{\hat{P}}_{#1}}}{{\mathbf{\hat{P}}_{#1}^{({#2})}}}
        }
}

\DeclareDocumentCommand\ccc{ g g }{%
        \IfNoValueTF {#1} {\mathbf{q}} {
            \IfNoValueTF {#2} {{\mathbf{q}_{#1}}}{{\mathbf{q}_{#1}^{(#2)}}{}}
        }
}

\DeclareDocumentCommand\constc{ g }{%
        \IfNoValueTF {#1} {c} {
            {c^{({#1})}}
        }
}

\DeclareDocumentCommand\setz{ g g }{%
        \IfNoValueTF {#1} {\mathcal{Z}} {
            \IfNoValueTF {#2} {\mathcal{Z}^{(#1)}}{\mathcal{Z}^{(#1)}_{#2}}
        }
}

\DeclareDocumentCommand\setzp{ g g }{%
        \IfNoValueTF {#1} {\mathcal{Z^+}} {
            \IfNoValueTF {#2} {\mathcal{Z}^{+(#1)}}{\mathcal{Z}^{+(#1)}_{#2}}
        }
}

\DeclareDocumentCommand\setzn{ g g }{%
        \IfNoValueTF {#1} {\mathcal{Z^-}} {
            \IfNoValueTF {#2} {\mathcal{Z}^{-(#1)}}{\mathcal{Z}^{-(#1)}_{#2}}
        }
}

\DeclareDocumentCommand\tsetz{ g g }{%
        \IfNoValueTF {#1} {\tilde{\mathcal{Z}}} {
            \IfNoValueTF {#2} {\tilde{\mathcal{Z}}^{(#1)}}{\tilde{\mathcal{Z}}^{(#1)}_{#2}}
        }
}

\DeclareDocumentCommand\tz{ g g }{%
        \IfNoValueTF {#1} {\tilde{z}} {
            \IfNoValueTF {#2} {\tilde{z}^{(#1)}}{\tilde{z}^{(#1)}_{#2}}
        }
}

\DeclareDocumentCommand\f{ g g }{%
        \IfNoValueTF {#1} {f} {
            \IfNoValueTF {#2} {f^{(#1)}}{f^{(#1)}_{#2}}
        }
}

\DeclareDocumentCommand\lf{ g g }{%
        \IfNoValueTF {#1} {\underline{f}} {
            \IfNoValueTF {#2} {\underline{f}^{(#1)}}{\underline{f}^{(#1)}_{#2}}
        }
}

\def\eqref#1{Eq.~(\ref{#1})}

\def\1{\bm{1}}

\def\vtheta{{\bm{\theta}}}

\def\vv{{\bm{v}}}

\def\vx{{\bm{x}}}

\def\vz{{\bm{z}}}

\DeclareMathAlphabet{\mathsfit}{\encodingdefault}{\sfdefault}{m}{sl}
\SetMathAlphabet{\mathsfit}{bold}{\encodingdefault}{\sfdefault}{bx}{n}

\begin{document}

\definecolor{mark}{rgb}{0.1, 0.5, 0.7}
\newcommand{\MARK}[1]{\textcolor{mark}{#1}}

\crefname{equation}{Eq.}{Eqs.}

\twocolumn[
\icmltitle{Are Diffusion Models Vulnerable to Membership Inference Attacks?}



\icmlsetsymbol{equal}{*}

\begin{icmlauthorlist}
\icmlauthor{Jinhao Duan}{drexel}
\icmlauthor{Fei Kong}{uestc}
\icmlauthor{Shiqi Wang}{amazon}
\icmlauthor{Xiaoshuang Shi}{uestc}
\icmlauthor{Kaidi Xu}{drexel}
\end{icmlauthorlist}


\icmlaffiliation{drexel}{Drexel University}
\icmlaffiliation{amazon}{AWS AI Lab}
\icmlaffiliation{uestc}{University of Electronic Science and Technology of China}

\icmlcorrespondingauthor{Kaidi Xu}{kx46@drexel.edu}

\icmlkeywords{Machine Learning, ICML}

\vskip 0.3in
]



\printAffiliationsAndNotice{}  

\begin{abstract}
Diffusion-based generative models have shown great potential for image synthesis, but there is a lack of research on the security and privacy risks they may pose. In this paper, we investigate the vulnerability of diffusion models to Membership Inference Attacks (MIAs), a common privacy concern. Our results indicate that existing MIAs designed for GANs or VAE are largely ineffective on diffusion models, either due to inapplicable scenarios (e.g., requiring the discriminator of GANs) or inappropriate assumptions (e.g., closer distances between synthetic samples and member samples).
To address this gap, we propose \textit{Step-wise Error Comparing Membership Inference} (SecMI), a query-based MIA that infers memberships by assessing the matching of forward process posterior estimation at each timestep.
SecMI follows the common overfitting assumption in MIA where member samples normally have smaller estimation errors, compared with hold-out samples.
We consider both the standard diffusion models, e.g., DDPM, and the text-to-image diffusion models, e.g., Latent Diffusion Models and Stable Diffusion.
Experimental results demonstrate that our methods precisely infer the membership with high confidence on both of the two scenarios across multiple different datasets.
Code is available at \url{https://github.com/jinhaoduan/SecMI}.
\end{abstract}


\section{Introduction}
Recently, diffusion models~\cite{song2019generative,song2020score,ho2020denoising} have dominated the image generation fields. Large-scale diffusion models, such as Stable Diffusion~\cite{rombach2022high}, DALLE-2~\cite{ramesh2022hierarchical}, Imagen~\cite{saharia2022photorealistic}, achieve remarkable fidelity and utility in text-to-image generation. Images generated by these models are full of diversity and creativity, which may profoundly change the way of human creation.

However, tremendous privacy risks~\cite{bommasani2021opportunities} and copyright disputes~\cite{ihalainen2018computer} are emerging, with the deployment of these generative models. For example, previous works demonstrate that GANs~\cite{goodfellow2020generative}/VAEs~\cite{kingma2013auto}-based generative models suffer from privacy leaking~\cite{hayes2017logan} and data reconstruction~\cite{zhang2020secret} issues, yet we are still unaware of whether diffusion models have similar concerns. On the other side, artists are getting unions against commercial large-scale generative models recently~\cite{bbc2022Art,cnn2022AI,washington2022AI}, since the creations of human artists may be exploited unauthorizedly.
Lots of security assessments and risk screening need to be addressed before releasing these models.

Membership Inference Attacks (MIAs)~\cite{cornellMI} is one of the most common privacy risks that associated with various privacy concerns.
For a given pre-trained model, MIAs aim to identify the membership of a single sample as \textit{member sample} that comes from the member set (training set), or as \textit{hold-out sample} that comes from the hold-out set.
Although MIAs have been well explored in the classic classification models and conventional generative models, related works on diffusion models are still missing in the literature.
In this paper, we intend to answer the following question:

\textit{Are diffusion-based generative models vulnerable to MIAs?}

We first summarize existing MIAs designed for conventional generative models and evaluate their performances in the diffusion model setting. Specifically, we summarize five MIAs and make them available to the diffusion models by either training shadow models~\cite{shokri2017membership} or providing the required information in an alternative manner.
Our results indicate that these methods are largely ineffective for diffusion models due to various potential reasons, such as more standard and stronger evaluations with larger member sets, limited exploitation of the specific properties of diffusion models, and inappropriate assumptions, i.e., synthetic images are closer to member images~\cite{hu2021membership, chen2020gan, mukherjee2021privgan}.
The lacking of effective MIAs may cause a false sense of security for diffusion models.

\begin{table*}
    \centering
    \caption{Taxonomy of MIAs against generative models over the previous works and our work. All the methods are evaluated on DDPM trained with a 50\% training split of the CIFAR-10 dataset as the member set and take the rest of the training split as the hold-out set. DMs stands for Diffusion Models.}
    \label{tab:Taxonomy}
     \vspace{-2mm}
    \adjustbox{width=1\textwidth}{
    
    \begin{threeparttable}
        \begin{tabular}{clccccc} \toprule
         Attack Type & \multicolumn{1}{l}{Method} & Discriminator & Generator & Synthetic & Applicable to DMs & ASR to DMs $\uparrow$ \\
         \cmidrule(lr){1-2} \cmidrule(lr){3-5} \cmidrule(lr){6-7}
         - & Random Guess & \XSolidBrush & \XSolidBrush & \XSolidBrush & YES & 0.500 \\
         \midrule
         White-box & GAN-Leaks (white-box)~\cite{chen2020gan} & \XSolidBrush & $\square$ & \XSolidBrush & NO & 0.615$\dagger$ \\
         \midrule
         \cmidrule(lr){2-7}
         \multirow{5}{*}{Black-box} & Shadow Model + LOGAN & \XSolidBrush & \XSolidBrush & \CheckmarkBold & YES & 0.544 \\
         & Shadow Model + TVD & \XSolidBrush & \XSolidBrush & \CheckmarkBold & YES & 0.089$^{\ddagger}$ \\
         & Over-Representation~\cite{hu2021membership} & \XSolidBrush & \XSolidBrush & \CheckmarkBold & YES & 0.532 \\
         & Monte-Carlo Set~\cite{hilprecht2019monte} & \XSolidBrush & \XSolidBrush & \CheckmarkBold & YES & 0.505 \\
         & GAN-Leaks (black-box)~\cite{chen2020gan} & \XSolidBrush & \XSolidBrush & \CheckmarkBold & YES & 0.507 \\
         \midrule
         \multirow{4}{*}{Query-based} & LOGAN~\cite{hayes2017logan} & $\blacksquare$ & \XSolidBrush & \XSolidBrush & NO & - \\
         & TVD~\cite{mukherjee2021privgan} & $\blacksquare$ & \XSolidBrush & \XSolidBrush & NO & - \\
         & SecMI$_{\textit{stat}}$ (ours) & \XSolidBrush & $\blacksquare$ & \XSolidBrush & YES & 0.811 \\
         & SecMI$_{\textit{NNs}}$ (ours) & \XSolidBrush & $\blacksquare$ & \XSolidBrush & YES &  0.888 \\
         
         \bottomrule
    \end{tabular}
    \begin{tablenotes}
        \centering
        \item \CheckmarkBold: with access $\,\,\,$ \XSolidBrush: without access $\,\,\,$ $\square$: white-box access $\,\,\,$ $\blacksquare$: black-box access
        
        \raggedright
        \item[$\dagger$]: GAN-leaks (white-box) is computationally infeasible for diffusion models. Here is the theoretical performance upper-bound of GAN-leaks, by generating the precise latent code of generated data through DDIM. \\
        \item[$\ddagger$]: \cite{mukherjee2021privgan} measures the attack efficiency by calculating the upper bound Total Variation Distance (TVD $\in$ [0, 1], a higher value means better attack performance). \\

    \end{tablenotes}
    \end{threeparttable}
    }
    \vspace{-3mm}
\end{table*}

To mitigate this gap, we consider designing MIAs by leveraging the specific properties of diffusion models.
Our motivation derives from the learning objectives where diffusion models are trained to match the forward process posterior distribution at each timestep with a parameterized model $\epsilon_{\theta}$~\cite{ho2020denoising}.
Recall that it is commonly known that membership privacy leaking benefits from the overfitting issue~\cite{shokri2017membership,privacyoverfittingmi}: member samples normally are memorized ``better'' than hold-out samples. Therefore a natural assumption regarding the membership exposure to the diffusion models is that member samples may have smaller posterior estimation errors compared with hold-out samples.

Based on that, we propose Step-wise Error Comparing Membership Inference (SecMI) to investigate the privacy leaking of diffusion models. SecMI is a query-based MIA that only relies on the inference results and can be applied to various diffusion models.
We study three popular diffusion models: DDPM~\cite{ho2020denoising}, Latent Diffusion Models (LDMs)~\cite{rombach2022high}, and Stable Diffusion, across multiple popular datasets, including CIFAR-10/100~\cite{krizhevsky2009learning}, STL10-Unlabeled (STL10-U)~\cite{coates2011analysis}, Tiny-ImageNet (Tiny-IN) for DDPM, Pokemon~\cite{pinkney2022pokemon} and COCO2017-val~\cite{lin2014microsoft} for LDMs, and Laion~\cite{schuhmann2022laion} for Stable Diffusion. Our contributions can be summarized as the following:
\begin{itemize}
  \setlength\itemsep{-1pt}

    \item To the best of our knowledge, this is the first work that investigates the vulnerability of diffusion models on MIAs. We summarize conventional MIAs and evaluate their performances on diffusion models. Results show that they are largely ineffective.
    \item We propose SecMI, a query-based MIA relying on the error comparison of the forward process posterior estimation. We apply SecMI on both the standard diffusion model, e.g., DDPM, and the state-of-the-art text-to-image diffusion model, e.g., Stable Diffusion. 
    \item We evaluate SecMI across multiple datasets and report attack performance, including the True-Positive Rate (TPR) at low False-Positive Rate (FPR)~\cite{carlini2022membership}. Experimental results show that SecMI precisely infers the membership across all the experiment settings ($\geq 0.80$ avg. Attack Success Rate (ASR) and $\geq 0.85$ avg. Area
Under Receiver Operating Characteristic (AUC)). 
\end{itemize}


\section{Related Works}

\paragraph{Generative Diffusion Models}
Different from Generative Adversarial Networks (GAN)~\cite{goodfellow2020generative,yuan2020attribute,yuan2023dde},
 diffusion models refer to specific latent variable models that approximate the real data distribution by matching a diffusion process with a parameterized reverse process~\cite{sohl2015deep,ho2020denoising}.
The diffusion process and reverse process can be either the continuous Langevin dynamics~\cite{song2019generative} or the discrete Markov chains~\cite{ho2020denoising}, which are proven to be equivalent to the variance-preserving (VP) SDE in~\cite{song2020score}.
Various diffusion models are proposed for speeding up inference~\cite{song2020denoising,salimans2022progressive,dockhorn2022scorebased,xiao2022tackling,watson2022learning,rombach2022high, meng2021sdedit}, conditional generation~\cite{dhariwal2021diffusion,ho2022classifier,meng2021sdedit}, and multi-modality generative tasks, such as audio synthesis~\cite{kong2020diffwave}.

\paragraph{Membership Inference Privacy}
Membership Inference Attack~\cite{cornellMI} has been widely explored for classification models. MIAs can be recognized as black-box attacks~\cite{shokri2017membership,ndss19salem,yeom2018privacy,sablayrolles2019white,song2020systematic,choquette2021label,hui2021practical,truex2019demystifying,salem2018mlleaks,locationmi} and white-box attacks~\cite{nasr2019comprehensive,rezaei2020towards}, depending on the accessibility to the target models.  \cite{choquette2021label} shows that logits are not necessary for MIA and proposes a label-only attack. \cite{carlini2022membership} reveals that MIA should be evaluated with strict metrics due to the fact that correctly and incorrectly inferring a membership are not equally important.
\cite{sablayrolles2019white} proves that 
black-box MIA can approximate white-box MIA performance under certain assumptions on the model weights distribution. In this paper, we mainly consider the query-based settings, i.e., only access to the query results of diffusion models.

Similarly, in terms of generative models, \cite{hayes2017logan} reveals that memberships can be effectively identified by the logits of the discriminator from GANs. 
 \cite{hilprecht2019monte} proposes the Monte Carlo score for black-box MIA, and the Reconstruction Attack for VAE by leveraging the reconstruction loss item. The Monte Carlo score measures the distance between synthetic samples and member samples, which is further adopted by several works~\cite{hu2021membership, chen2020gan, mukherjee2021privgan}.

Concurrently, \cite{hu2023membership,wu2022membership,carlini2023extracting} also investigate the MIA issues in diffusion models. \cite{wu2022membership} assumes that the member set and the hold-out set come from different distributions, which makes their MIAs much easier. Our paper follows the common ``MI security game protocol''~\cite{hu2023membership,carlini2023extracting} where the member set and the hold-out set are in the same distribution.
\cite{hu2023membership} infers membership by comparing the loss values of member samples and hold-out samples. 
\cite{carlini2023extracting} discloses that training data extraction and MIAs are both feasible for diffusion models. They share a similar loss-based MIA idea with~\cite{hu2023membership} and further incorporate with the powerful LiRA~\cite{carlini2022membership} to improve the attack performance.
Different from that, our method identifies membership by assessing the posterior estimation with deterministic sampling and reversing, which achieves much stronger performances with the simple threshold inference strategy.


\section{Preliminary Analysis}
In this section, we formally define the membership inference problem and investigate whether existing MIAs designed for GANs or VAE work for diffusion models.

\subsection{Problem Statement}
\textit{Membership Inference} (MI) aims to predict whether or not a specific sample is used as a training sample.
Given a model $f_{\theta}$ parameterized by weights $\theta$ and dataset $D=\{\vx_1, \cdots, \vx_n\}$ drawn from data distribution $q_{\textit{data}}$, we follow the common assumption~\cite{sablayrolles2019white,carlini2022membership} where $D$ is split into two subsets, $D_{\textit{M}}$ and $D_{\textit{H}}$, and $D = D_{\textit{M}} \cup D_{\textit{H}}$. $f_{\theta}$ is solely trained on $D_{\textit{M}}$.
In this case, $D_{\textit{M}}$ is the \textit{member set} of $f_{\theta}$ and $D_{\textit{H}}$ is the \textit{hold-out set}.
Each sample $\vx_i$ is equipped with a membership identifier $m_i$, where $m_i = 1$ if $\vx_i \sim D_{\textit{M}}$; otherwise $m_i = 0$.
The attacker only has access to $D$ while having no knowledge about $D_{\textit{M}}$ and $D_{\textit{H}}$.
An attack algorithm $\mathcal{M}$ is designed to predict whether or not $\vx_i$ is in $D_{\textit{M}}$:
\begin{figure}[t]
    \centering
    \includegraphics[width=0.48\textwidth]{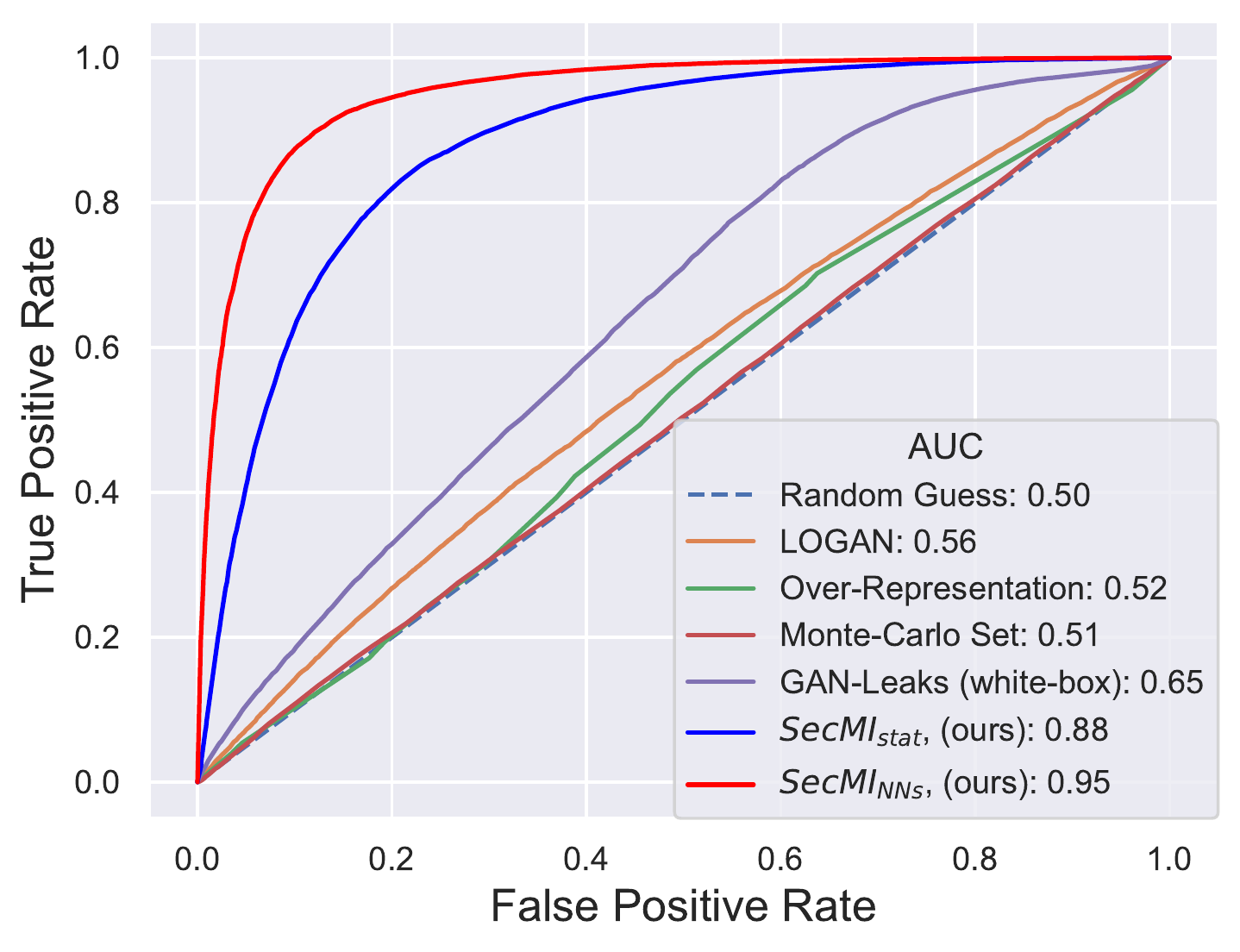}
    \caption{Comparing the TPR v.s. FPR of prior MIAs designed for generative models. Evaluations are conducted on DDPM with half of the CIFAR-10 training split as the member set and the other half as the hold-out set. Prior MIAs are largely ineffective on DDPM.}
    \label{fig:existing_MIAs_ROC}
\end{figure}
\begin{equation}
    \mathcal{M}(\vx_i, \theta) = \mathds{1}\left[ \mathds{P}(m_i = 1|\theta, \vx_i) \geq \tau \right]
\end{equation}
where $\mathcal{M}(\vx_i, \theta) = 1$ means $x_i$ comes from $D_{\textit{M}}$,  $\mathds{1}{\left[ A \right]} = 1$ if $A$ is true, and $\tau$ is the threshold.
For the generative model scenario, we reuse $\theta$ as the weights of generator $G$ and let $p_{\theta}(\vx)$ denote the generative distribution where generated sample $\vx \sim p_{\theta}(\vx|\vz)$ given latent code $\vz$.

\textbf{Evaluation Metrics.}
Following the most convincing metrics used in MIAs~\cite{carlini2022membership, choquette2021label}, we measure the performance of MIAs with Attack Success Rate (ASR), Area Under Receiver Operating Characteristic (AUC), and True-Positive Rate (TPR) at extremely low False-Positive Rate (FPR), e.g., TPR@1\%/0.1\% FPR.

\subsection{Evaluating Existing MIAs on Diffusion Models}\label{sec:existing_mia}
As summarized in~\cref{tab:Taxonomy}, we consider five different MIAs designed for generative models: LOGAN~\cite{hayes2017logan}, TVD~\cite{mukherjee2021privgan}, Over-Representation~\cite{hu2021membership}, Monte-Carlo Set~\cite{hilprecht2019monte}, and GAN-Leaks~\cite{chen2020gan}, as summarized in~\cref{tab:Taxonomy}.
Since LOGAN and TVD require access to the discriminator of GANs, we train Shadow Models~\cite{shokri2017membership} to make them available to diffusion models.
All the methods are evaluated on DDPM~\cite{ho2020denoising} trained over CIFAR-10~\cite{krizhevsky2009learning} with 50\% training split as the member set and the rest as the hold-out set. Also, we choose StyleGAN2~\cite{karras2020analyzing} as the Shadow Model. 

The ROC curves of these methods are presented in~\cref{fig:existing_MIAs_ROC}. 
For black-box MIAs that only require the query results of the generator, only LOGAN shows marginal effectiveness on diffusion models while other MIAs are largely ineffective. The original white-box GAN-leaks requires the full gradient to optimize the latent code, which is computationally infeasible for diffusion models. Here we calculate its theoretical upper bound by providing the latent codes generated by DDIM~\cite{song2020denoising}.
In this setting, the white-box GAN-leaks shows certain effectiveness.

\subsection{Analytical Insights}\label{sec:reason_failed}
We summarize the potential reasons for the limited success of existing methods in diffusion models:

\textbf{Stronger Evaluations.} Previous works~\cite{hayes2017logan, chen2020gan, hilprecht2019monte,hu2023membership} adopt very limited member set size (e.g., $\leq$ 10\% of the training split of $D$) while we employ a size of 50\%. It is known that MIAs benefit from overfitting~\cite{shokri2017membership,privacyoverfittingmi}. Smaller member set may exacerbate overfitting, which amplifies the effect of these methods.

\textbf{Diffusion Models Generalize Better.}
Prior works mainly assume that a sample that ``occurred'' in a higher frequency when sampling from the generator, is more likely to be in the member set:
\begin{equation}
\label{eq:exist_assumption}
    \mathds{P}(m_i=1|\theta_G, \vx_i) \propto \mathds{P}(\vx_i|\theta_G).
\end{equation}
However, this holds when the generative distribution overfits the member set, i.e., $d(p_\theta, p_{\textit{D}_\textit{M}}) < d(p_{\theta}, p_{\textit{D}_\textit{H}})$ where $d(\cdot, \cdot)$ is a distance measurement and $p_\theta, p_{\textit{D}_\textit{M}}, p_{\textit{D}_\textit{H}}$ are generative distribution, member distribution, and hold-out distribution, respectively.
To measure it, we estimate $d(p_\theta, p_{\textit{D}_\textit{M}})$ and $d(p_{\theta}, p_{\textit{D}_\textit{H}})$ by calculating the FIDs~\cite{heusel2017gans} between 25,000 synthetic images and 25,000 member/hold-out samples, and we get 9.66 v.s. 9.85, which shows diffusion models have no distinct bias toward the member samples. 

\textbf{Limited Exploitation of Diffusion Models.} Existing MIAs were primarily designed for GANs or VAEs, and thus do not take into account the specific properties of diffusion models. 


\section{Methodology}

In this section, we provide the first MIA design for diffusion models exploiting the step-wise forward process posterior estimation.

\subsection{Notations}
We follow the common notations of diffusion models~\cite{ho2020denoising} where we denote by $q(\vx_0)$ the real data distribution and $p_\theta(\vx_0)$ the latent variable model approximating $q(\vx_0)$ with noise-prediction model $\epsilon_\theta$ parameterized by weights $\theta$.
Diffusion models consist of the $T$-step diffusion process $q(\vx_{t}|\vx_{t-1})$ and the denoising process $p_{\theta}(\vx_{t-1}|\vx_{t}),  (1 \leq t \leq T)$, with the following transitions:
\begin{equation}
\begin{aligned}
q(\vx_{t}|\vx_{t-1}) = \mathcal{N}(\vx_{t}; \sqrt{1 - \beta_t}\vx_{t-1},  \beta_t \textbf{I}) \,\,\, \,\,\,\\
    p_{\theta}(\vx_{t-1}|\vx_{t}) = \mathcal{N}(\vx_{t-1};\mu_{\theta}(\vx_t, t),  \Sigma_{\theta}(\vx_t, t))
\end{aligned}
\end{equation}
where $\beta_1, \cdots, \beta_{T}$ is a variance schedule. The forward sampling at arbitrary time step $t$ can be obtained by
\begin{equation}
    q(\vx_t | \vx_0) = \mathcal{N}(\vx_t;\sqrt{\bar{\alpha}_t}\vx_0, (1 - \bar{\alpha}_t) \textbf{I}),
\end{equation}
where $\alpha_t = 1 - \beta_t$ and $\bar{\alpha}_t=\prod_{s=1}^t{\alpha_s}$.

\subsection{Exposing Membership via Step-Wise Error Comparison}\label{sec:exposing_membership}

\cite{sablayrolles2019white} demonstrates that the Bayes optimal performance of MIA can be approximated as
\begin{equation}
\label{eq:opt_black_mia}
    \mathcal{M}_{opt}(\vx, \theta) = \mathds{1}\left[ \ell(\theta, \vx) \leq \tau \right],
\end{equation}
under a mild assumption on the model weights distribution,
where $\tau$ is the threshold and $\ell$ is the loss function, i.e., the cross-entropy loss for classification tasks. 
\cref{eq:opt_black_mia} reveals that membership can be exposed by measuring how well $\theta$ is learned on data point $\vx$.

For diffusion models, a similar metric can also be derived from the learning objective.
Recall that diffusion models are trained to optimize the variational bound $p_{\theta}(\vx_0)$ by matching the forward process posteriors at each step $t$:
\begin{equation}
\label{eq:diffusion_model_loss}
    \ell_t = \mathbb{E}_{q} \left [ \frac{1}{2\sigma_{t}^2} || \tilde{\mu}_t (\vx_t, \vx_0) - \mu_{\theta}(\vx_t, t) ||^2 \right],
\end{equation}
where $\tilde{\mu}_t (\vx_t, \vx_0)$ is the mean of the posterior distribution $q(\vx_{t-1}|\vx_t, \vx_0)$ and $\mu_{\theta}(\vx_t, t)$ refers to the estimation.
$||\cdot||^2$ refers to the mean squared error (MSE).
\cref{eq:diffusion_model_loss} indicates that the local estimation error of single data point $\vx_0$ at timestep $t$ is
\begin{equation}\label{eq:sample_wise_err}
\ell_{t, \vx_0} = || \hat{\vx}_{t-1} - \vx_{t-1} ||^2,    
\end{equation}
where $\vx_{t-1} \sim q(\vx_{t-1} | \vx_{t}, \vx_0)$ and $\hat{\vx}_{t-1} \sim p_{\theta}(\hat{\vx}_{t-1}|\vx_{t})$ (we omit $\frac{1}{2\sigma_t^2}$ since it is constant).
A natural assumption regarding the membership exposure is that samples from the member set $D_{\textit{M}}$ may have smaller estimation errors at step $t$, compared with samples from the hold-out set $D_{\textit{H}}$:
\begin{equation}
\label{eq:reconstruction_original}
     \ell_{t, \vx_{m}} \leq \ell_{t, \vx_{h}} , \,\, 1 \leq t \leq T,
\end{equation}
where $\vx_{m} \sim D_{\textit{M}}$ and $\vx_{h} \sim D_{\textit{H}}$.

However, the above quantity is intractable since it involves $q(\vx_{t-1} | \vx_{t}, \vx_0)$ and $p_{\theta}(\hat{\vx}_{t-1}|\vx_{t})$. Due to the essence of non-deterministic diffusion and denoising processes, i.e., the Markov Chain, it is intractable to calculate their analytical solutions.
Although estimating it with Monte Carlo sampling is possible, it is time-consuming for diffusion models requiring multi-step denoising, e.g., $T$ = 1,000.
Inspired by recent works on deterministic reversing and sampling from diffusion models~\cite{song2020denoising, kim2022diffusionclip,  song2020score}, we consider approximate~\cref{eq:sample_wise_err} with deterministic processes:
\begin{equation}
\begin{aligned}
    \vx_{t+1} & = \phi_{\theta}(\vx_t, t) \\
    &= \sqrt{\bar{\alpha}_{t+1}}f_\theta(\vx_t, t) + \sqrt{1 - \bar{\alpha}_{t+1}}\epsilon_{\theta}(\vx_t, t),
\end{aligned}
\end{equation}
\begin{equation}
\begin{aligned}
    \vx_{t-1} & = \psi_{\theta}(\vx_t, t) \\
    & = \sqrt{\bar{\alpha}_{t-1}}f_\theta(\vx_t, t) + \sqrt{1 - \bar{\alpha}_{t-1}}\epsilon_{\theta}(\vx_t, t),
\end{aligned}
\end{equation}
where
\begin{equation}
    f_\theta(\vx_t, t) = \frac{\vx_t - \sqrt{1 - \bar{\alpha}_t}\epsilon_\theta(\vx_t, t)}{\sqrt{\bar{\alpha}_t}}.
\end{equation}
We denote by $\Phi_{\theta}(x_s, t)$ the deterministic reverse, i.e., from $x_s$ to $x_t$ ($s < t$), and $\Psi_{\theta}(x_t, s)$ the deterministic denoise process, i.e., from $x_t$ to $x_s$:
\begin{equation}
\begin{aligned}
    \vx_{t} & = \Phi_{\theta}(\vx_s, t) = \phi_{\theta}( \cdots \phi_{\theta}( \phi_{\theta}(\vx_s, s), s+1), t-1) \\
    \vx_{s} & = \Psi_{\theta}(\vx_t, s) = \psi_{\theta}( \cdots \psi_{\theta}( \psi_{\theta}(\vx_t, t), t-1), s+1)
\end{aligned}
\end{equation}

\begin{figure}[ht]
    \centering
    \includegraphics[width=0.45\textwidth]{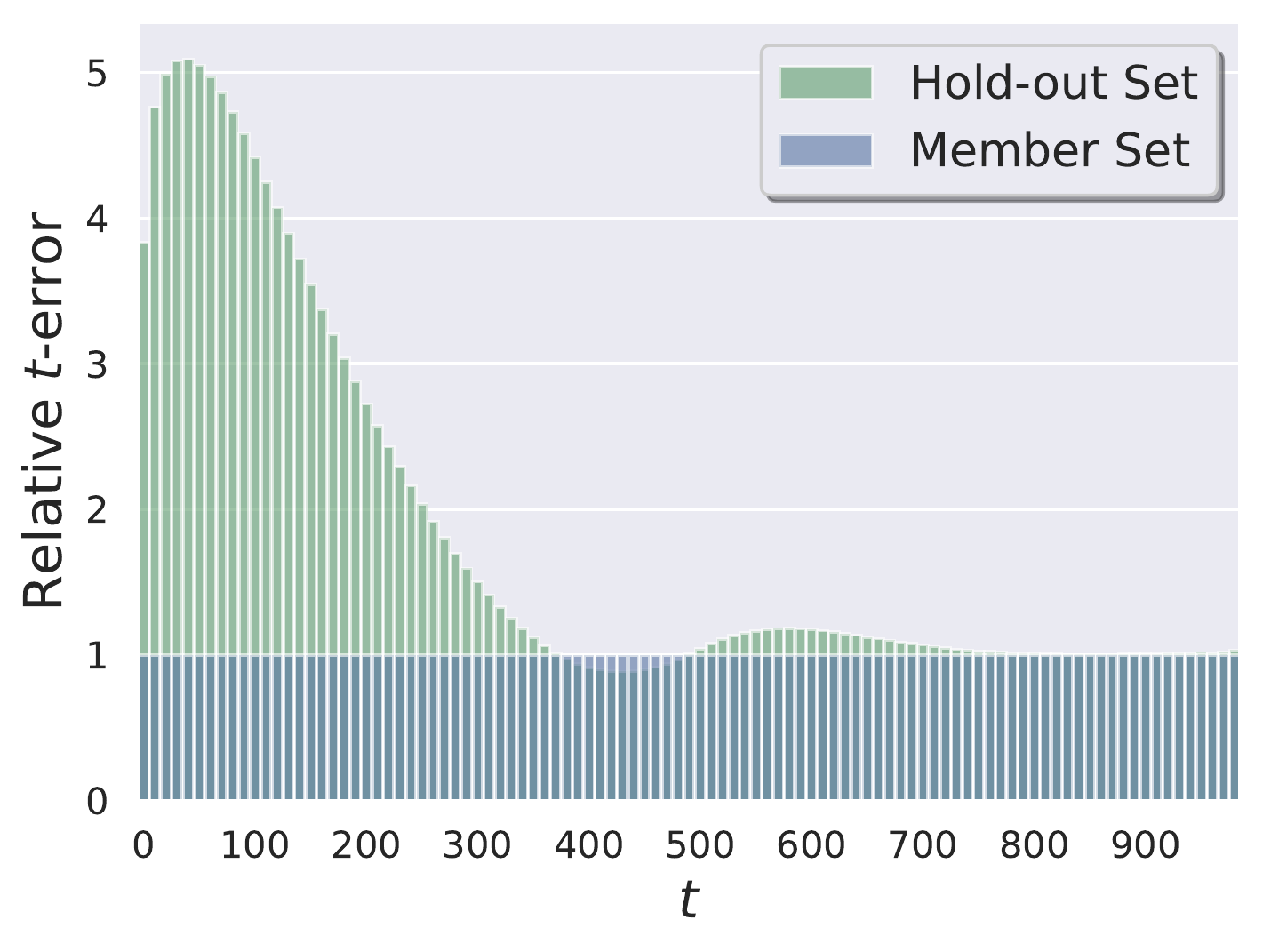}
     \vspace{-2mm}
    \caption{Comparisons of $t$-\textit{errors} for samples from the member and hold-out sets. Since the magnitudes are different at each step $t$, we set the $t$-\textit{errors} of member set as 1 at every timestep and report the relative sizes of $t$-\textit{errors} of Hold-out Set. It is shown that samples from the hold-out set have higher $t$-\textit{errors} compared with samples from the member set, proving that $t$-\textit{error} is an effective metric for identifying memberships.}
    \label{fig:errors_per_timestep}
     \vspace{-2mm}
\end{figure}

Then, we define $t$-\textit{error} as the approximated posterior estimation error at step $t$.
\begin{definition}[$t$-\textit{error}]\label{def:t_error}
For given sample $\vx_0 \sim D$ and the deterministic reverse result $\tilde{\vx}_t = \Phi_{\theta}(\vx_0, t)$ at timestep $t$, the approximated posterior estimation error at step $t$ is defined as $t$-\textit{error}:
\begin{equation}\label{eq:t_error}
    \tilde{\ell}_{t, \vx_0} = || \psi_{\theta} ( \phi_{\theta}(\tilde{\vx}_t, t), t) - \tilde{\vx}_{t} ||^2.
\end{equation}
\end{definition} 

We prove that $\tilde{\ell}_{t, \vx_0}$ is converged to $\ell_{t, \vx_0}$ when the diffusion model is converged to the learning objective, i.e., $\tilde{\ell}_{t, \vx_0} \rightarrow \ell_{t, \vx_0}$, when $|| \epsilon_{\theta}(\vx_t, t) - \epsilon ||^2 \rightarrow 0$ at any timestep $t$, where $\epsilon \sim \mathcal{N}(0, {\textbf{I}})$ (the ``noise-prediction'' loss in~\cite{ho2020denoising}:
\begin{equation}
\begin{aligned}
    \Delta_{t, \vx_0} & = \ell_{t, \vx_0} - \tilde{\ell}_{t, \vx_0} \\
    & = || \epsilon_{\theta}(\vx_t, t) - \epsilon ||^2 \\
    & - ||\sqrt{1 - \bar{\alpha}_t}(\epsilon_{\theta}(\tilde{\vx}_t, t) - \epsilon_{\theta}(\phi_{\theta}(\tilde{\vx}_t, t), t+1)) ||^2.
\end{aligned}
\end{equation}
It is shown that $\Delta_{t, \vx_0} \rightarrow 0$ when $|| \epsilon_{\theta}(\vx_t, t) - \epsilon ||^2 \rightarrow 0$ at any timestep $t$.

\cref{fig:errors_per_timestep} presents the relative scales of $t$-\textit{error} at each timestep for $D_{\textit{M}}$ and $D_{\textit{H}}$.
We show that samples from the hold-out set normally have larger $t$-\textit{errors} compared with member samples, which verifies the membership exposure assumption in~\cref{eq:reconstruction_original}. Besides, this phenomenon is getting distinct as $t$ approaches $0$. One of the possible reasons is that
$\vx_T$ is an approximated Gaussian Noise that contains almost zero information about the member sample while $\vx_0$ is the member sample.
A smaller $t$ triggers more ``memory'' about member samples, which results in serious privacy leaking.

\begin{table*}[ht]
    \centering
    \caption{Performance of SecMI on DDPM across four datasets. It is shown that both of the two variants achieve significant performances in all the settings.}
    \label{tab:ddpm_4ds_auc_asr}
     \vspace{-2mm}
    \resizebox{2\columnwidth}{!}{
    \begin{threeparttable}
    \begin{tabular}{ccccccccccccc}
        \toprule
         & & & \multicolumn{2}{c}{CIFAR-10} & \multicolumn{2}{c}{CIFAR-100} & \multicolumn{2}{c}{STL10-U} & \multicolumn{2}{c}{Tiny-IN} & \multicolumn{2}{c}{Average}\\
         \cmidrule(lr){4-5}
         \cmidrule(lr){6-7}
         \cmidrule(lr){8-9}
         \cmidrule(lr){10-11}
         \cmidrule(lr){12-13}
         Method & $\theta$ & \# Query & ASR$\uparrow$  & AUC$\uparrow$  & ASR$\uparrow$  & AUC$\uparrow$  & ASR$\uparrow$  & AUC$\uparrow$  & ASR$\uparrow$  & AUC$\uparrow$  & ASR$\uparrow$  & AUC$\uparrow$  \\
         
         \midrule
         
         GAN-Leaks$^\dagger$ & $\square$ & $\geq 1000 \times 2$ & 0.615 & 0.646 & 0.513 & 0.459 & 0.566 & 0.535 & 0.545 & 0.457 & 0.560 & 0.524 \\
         \midrule
         SecMI$_{stat}$ & $\blacksquare$ & $10 + 2$ & 0.811 & 0.881 & 0.798 & 0.868 & 0.809 & 0.881 & 0.821 & 0.894 & 0.810 & 0.881 \\
         SecMI$_{\textit{NNs}}$ $^\ddagger$ & $\blacksquare$ & $10 + 2$ & 0.888 & 0.951 & 0.872 & 0.940 & 0.892 & 0.950 & 0.903 & 0.956 & 0.889 & 0.949\\
         \bottomrule
    \end{tabular}
    \begin{tablenotes}
        \centering
        \item  $\square$: white-box access $\,\,\,$ $\blacksquare$: black-box access \\
        \raggedright
        \item $\dagger$: GAN-Leaks is computationally infeasible for diffusion models. Here are the theoretical results by providing the exact latent codes.\\
        \item $\ddagger$: We are aware that it is unfair to directly compare SecMI$_{\textit{NNs}}$ with other methods since it is evaluated on a slightly smaller member set and hold-out set. Here just to show the effectiveness of our method.
    \end{tablenotes}
    \end{threeparttable}
    }
\end{table*}

\begin{figure*}[ht]
        \begin{subfigure}[b]{0.25\textwidth}
                \includegraphics[width=\linewidth, height=4cm, trim=4 4 4 4,clip]{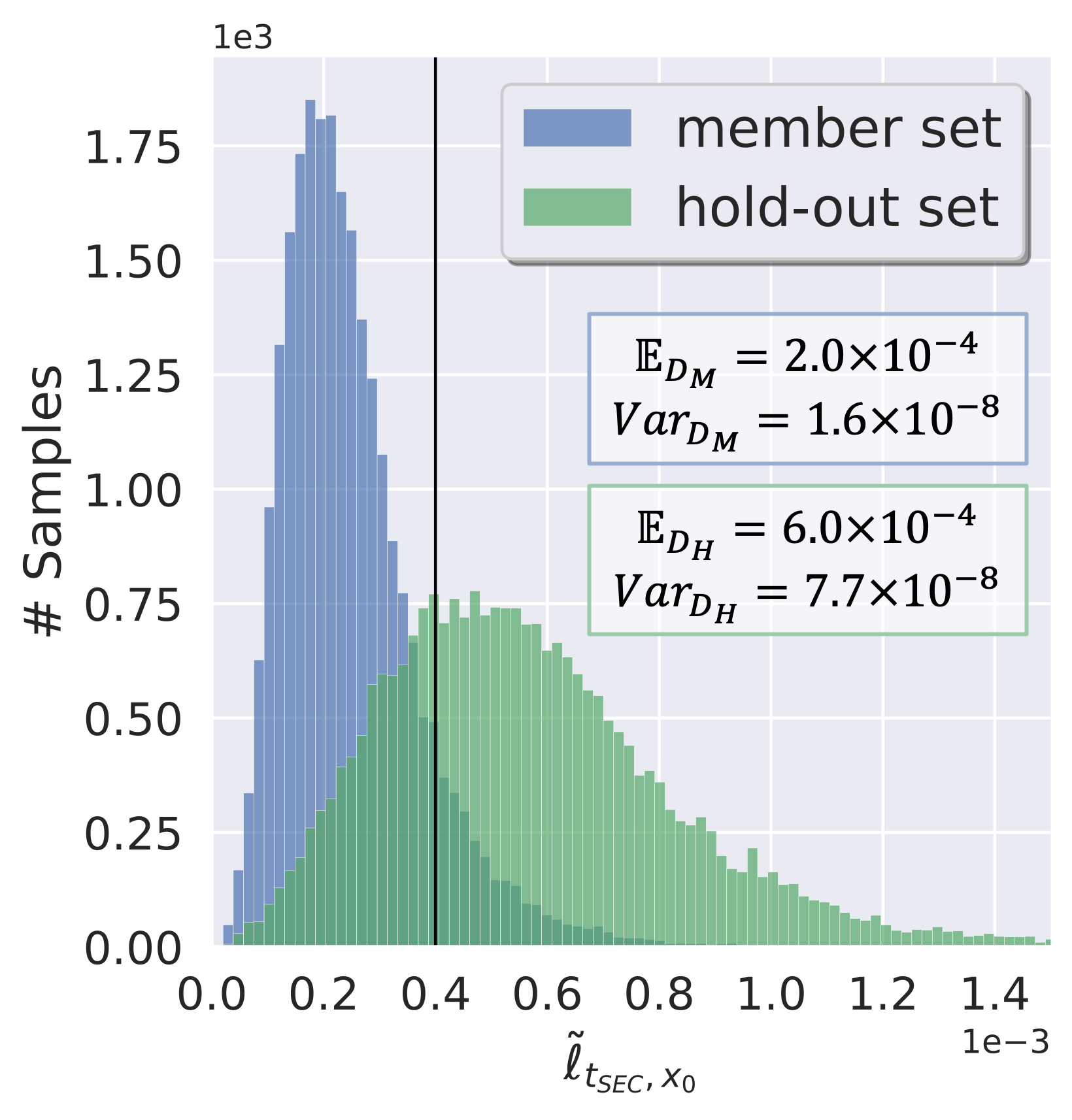}
                \caption{SecMI$_{stat}$ on CIFAR10.}
        \end{subfigure}%
        \begin{subfigure}[b]{0.25\textwidth}
                \includegraphics[width=\linewidth, height=4cm, trim=4 4 4 4,clip]{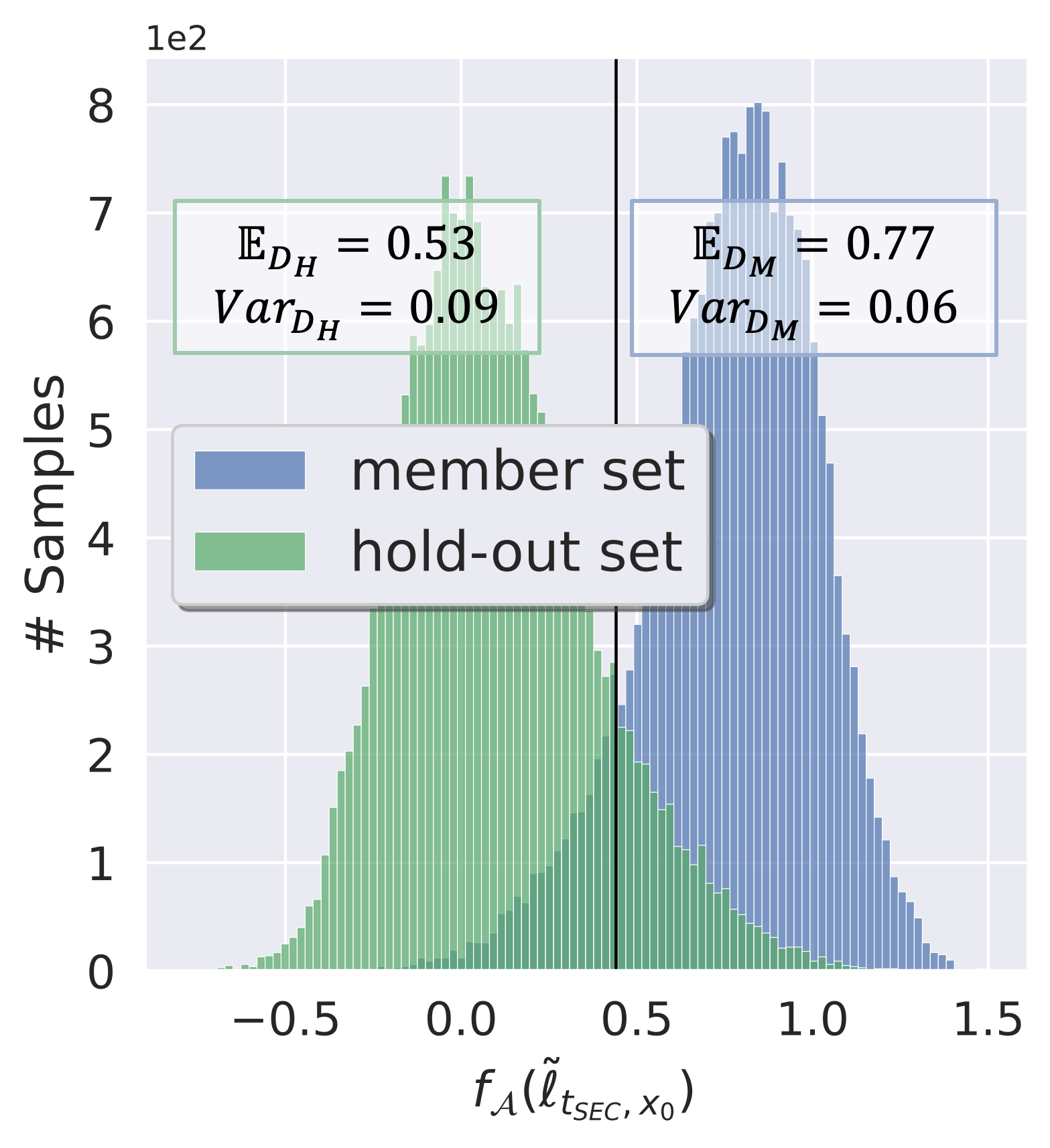}
                \caption{SecMI$_{\textit{NNs}}$ on CIFAR10.}
        \end{subfigure}%
        \begin{subfigure}[b]{0.25\textwidth}
                \includegraphics[width=\linewidth, height=4cm, trim=4 4 4 4,clip]{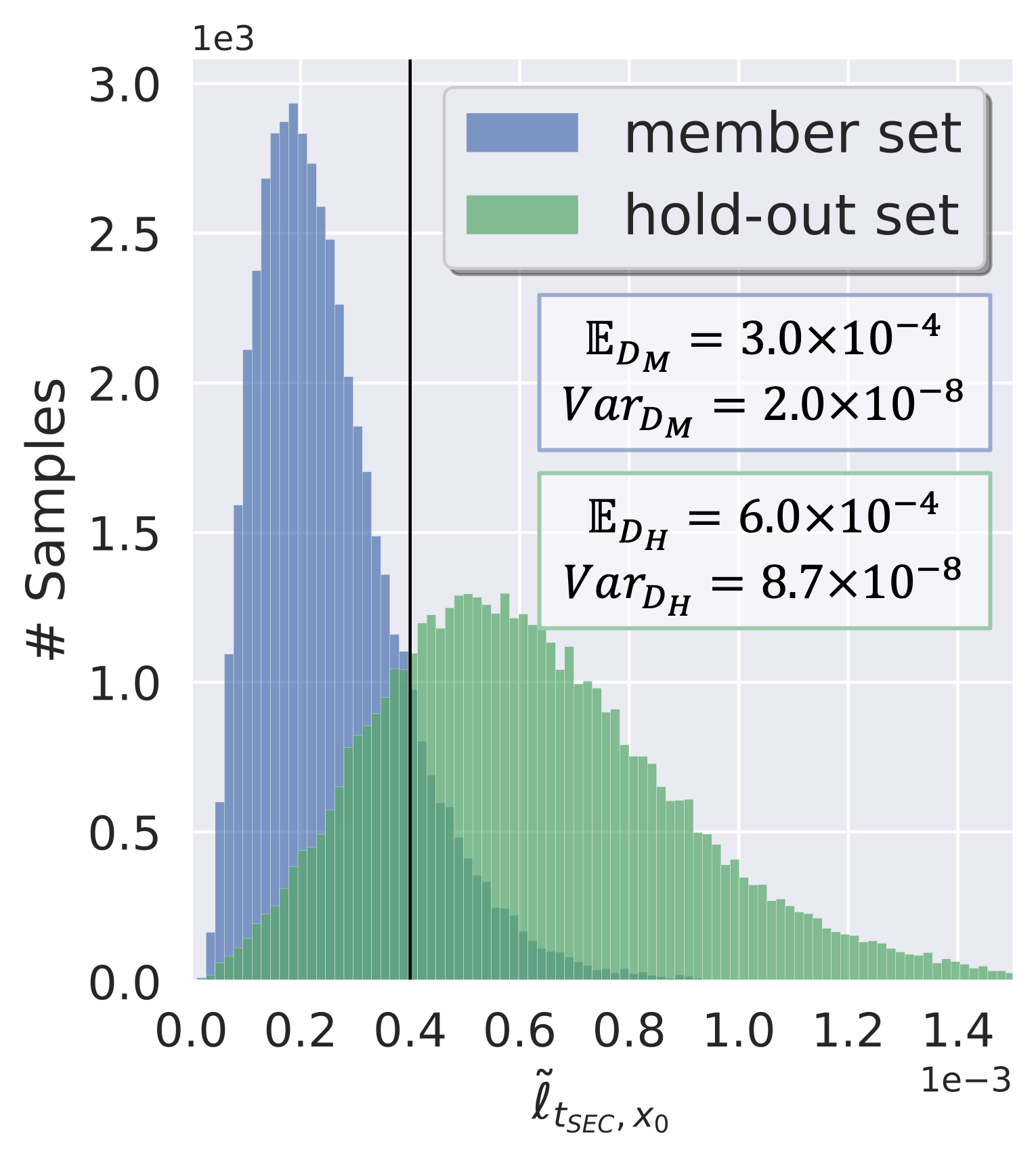}
                \caption{SecMI$_{stat}$ on Tiny-IN.}
        \end{subfigure}%
        \begin{subfigure}[b]{0.25\textwidth}
                \includegraphics[width=\linewidth, height=4cm, trim=4 4 4 4,clip]{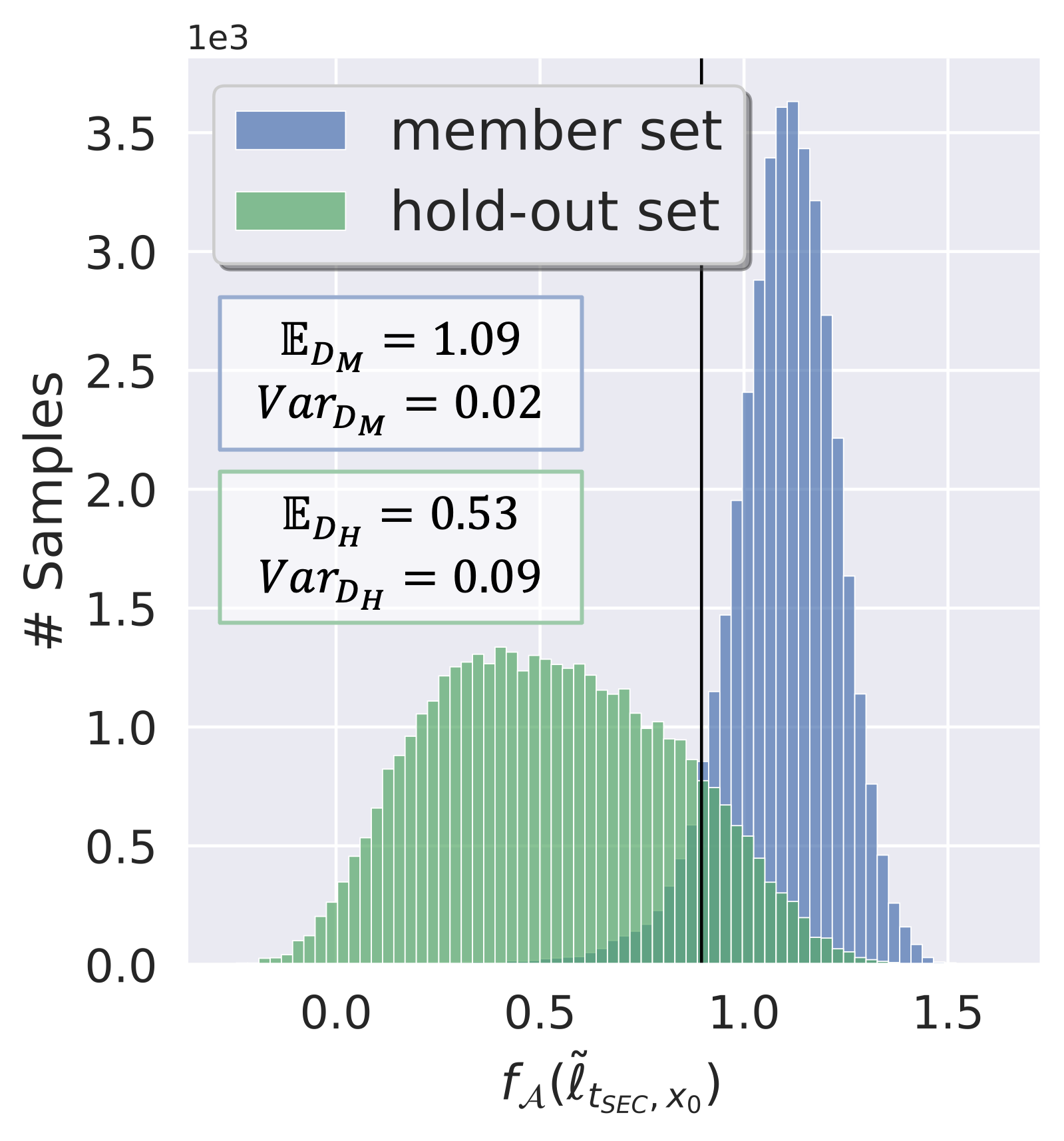}
                \caption{SecMI$_{\textit{NNs}}$ on Tiny-IN.}
        \end{subfigure}
        \caption{The $\tilde{\ell}_{t_{\textsc{SEC}}, x_0}$ and $f_{\mathcal{A}}(\tilde{\ell}_{t_{\textsc{SEC}}, x_0})$ distributions for samples from member set and hold-out set. It is clear that $t$-\textit{error} is a desirable indicator for membership identification. The vertical black line refers to the selected threshold $\tau$ for each figure.}\label{fig:secmi_dists}
         \vspace{-2mm}
\end{figure*}

\subsection{Error Comparing Based Membership Inference}
Our analysis indicates that comparing the approximated step-wise posterior estimation error, i.e., $t$-\textit{error}, is effective for identifying memberships. In this section, we design two strategies to infer membership by leveraging $t$-\textit{error}: the statistic-based inference, SecMI$_{stat}$, and the neural networks-based inference, SecMI$_{\textit{NNs}}$. 

We denote by $t_{\textsc{SEC}}$ the selected timestep for error comparing. For each sample $\vx_0 \sim D$, we calculate the corresponding $t$-\textit{error}, $\tilde\ell_{t_{\textsc{SEC}}, \vx_0}$, based on~\cref{eq:t_error}.
For SecMI$_{stat}$, we predict membership as the following:
\begin{equation}
\label{eq:SecMI_tau}
    \mathcal{M}(x_0, \theta) = \mathds{1}\left[ \tilde\ell_{t_{\textsc{SEC}}, \vx_0} \leq \tau \right],
\end{equation}
where $\tau$ is the threshold.
For SecMI$_{\textit{NNs}}$, we adopt an attack model $f_\mathcal{A}$ to infer membership. Specifically, $f_{\mathcal{A}}$ takes the pixel-wise absolute value of estimation error as the input and predicts the confidence of being a member sample. We use a 1-output classification model as the backbone of $f_{\mathcal{A}}$ and train it in a binary classification manner. We randomly sample a small subset of $D_{\textit{M}}$ and $D_{\textit{H}}$ as its training data. The membership is predicted as the following:
\begin{equation}
    \label{eq:SecMI_NNs}
    \mathcal{M}(\vx_0, \theta) = \mathds{1}\left[ f_{\mathcal{A}}( | \psi_{\theta} ( \phi_{\theta}(\tilde{\vx}_t, t), t) - \tilde{\vx}_{t} | ) \leq \tau \right].
\end{equation}
It is worth noting that although we only adopt the estimation error at a single timestep when predicting membership, experimental results show that this is already effective.
We may leave how to fuse multi-step errors for better membership inference in the future.

\subsection{Generalization to Various Diffusion Models}\label{sec:acclicability}
SecMI can be easily adapted to other popular diffusion models. To prove that, we provide the adaption to Latent Diffusion Models.
We leave the implementation details in~\cref{appendix:generalize_various_diffusion}.


\section{Experiments}

In this section, we evaluate the performance of SecMI across various datasets and settings.

\subsection{Experimental Setup}

\textbf{Datasets and Diffusion Models.}
For all  the datasets, we randomly select 50\% of the training samples as $D_{\textit{M}}$ and use the rest of the training samples as $D_{\textit{H}}$. For example, CIFAR-10 contains 50,000 images in the training set, so we have 25,000 images for $D_{\textit{M}}$ and another 25,000 images for $D_{\textit{H}}$. We summarize the adopted diffusion models and datasets in~\cref{appendix:adopted_model_datasets}.
In addition to the text-to-image settings, we adopt the image captions provided by the dataset organizers as the prompts.

\begin{table*}[ht]
    \centering
    \caption{The TPR at low FPR of SecMI on DDPM over four datasets.}
    \label{tab:ddpm_4ds_tpr}
    \vspace{-2mm}
    \resizebox{1.8\columnwidth}{!}{
    \begin{tabular}{ccccccccc}
        \toprule
         & \multicolumn{4}{c}{TPR @ 1\% FPR (\%) $\uparrow$ } & \multicolumn{4}{c}{TPR @ 0.1\% FPR (\%) $\uparrow$ } \\
         \cmidrule(lr){2-5}
         \cmidrule(lr){6-9}
         Methods & CIFAR-10 & CIFAR-100 & STL10-U & Tiny-IN & CIFAR-10 & CIFAR-100 & STL10-U & Tiny-IN \\
         \midrule
         GAN-Leaks & 2.80 & 1.85 & 1.17 & 1.01 & 0.29 & 0.23 & 0.24 & 0.13 \\
         \midrule   
         SecMI$_{stat}$ & 9.11 & 9.26 & 10.87 & 12.67 & 0.66 & 0.46 & 0.73 & 0.69 \\
         SecMI$_{\textit{NNs}}$ & 37.98 & 30.17 & 26.66 & 29.77 & 7.59 & 5.09 & 3.76 & 3.50 \\
         \bottomrule
    \end{tabular}
    }
\end{table*}
\begin{figure*}[ht]
\centering
        \begin{subfigure}[b]{0.25\textwidth}
                \includegraphics[width=\linewidth, height=3.5cm]{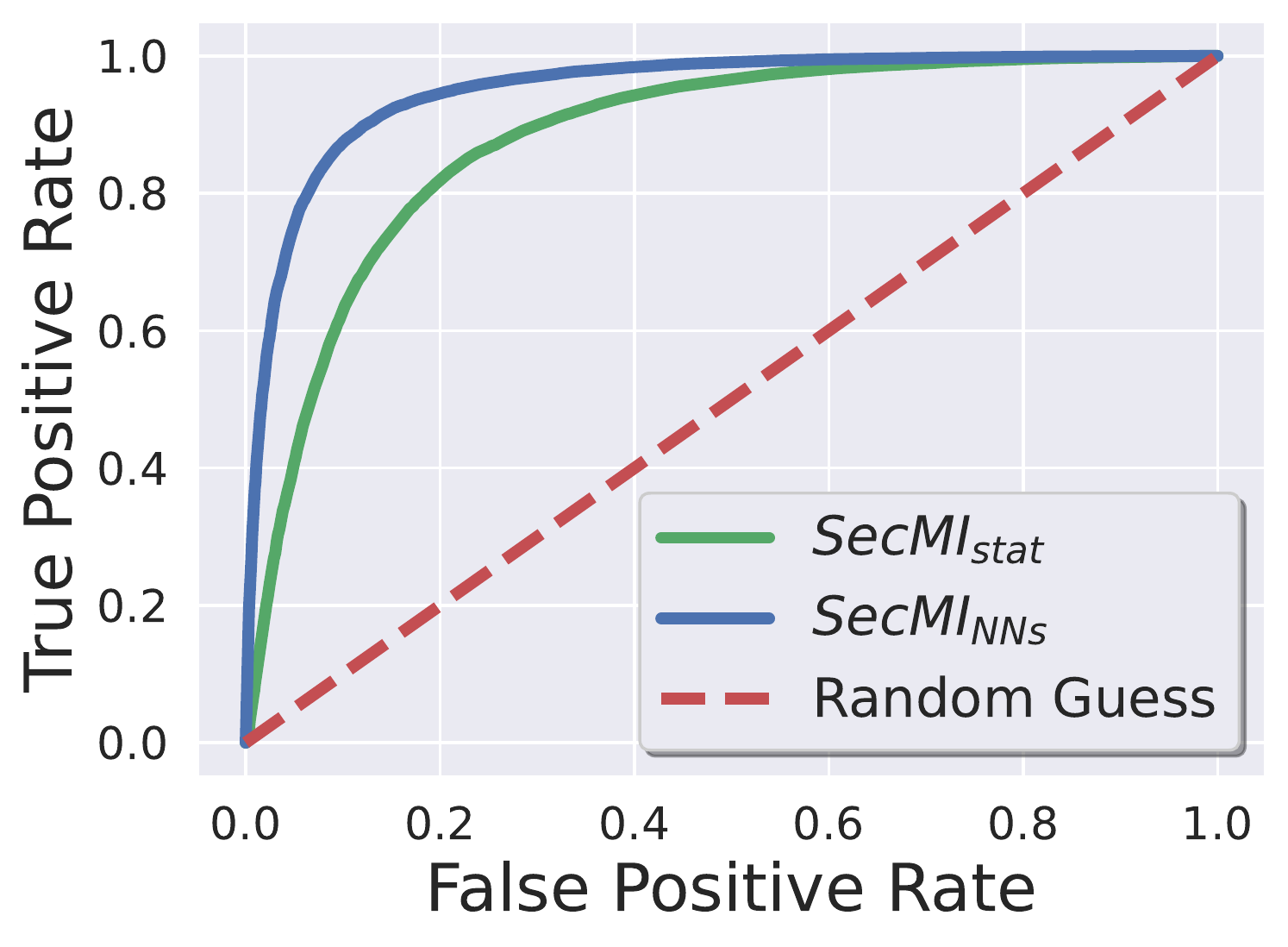}
                \caption{ROC on CIFAR10.}
        \end{subfigure}%
        \begin{subfigure}[b]{0.25\textwidth}
                \includegraphics[width=\linewidth, height=3.5cm]{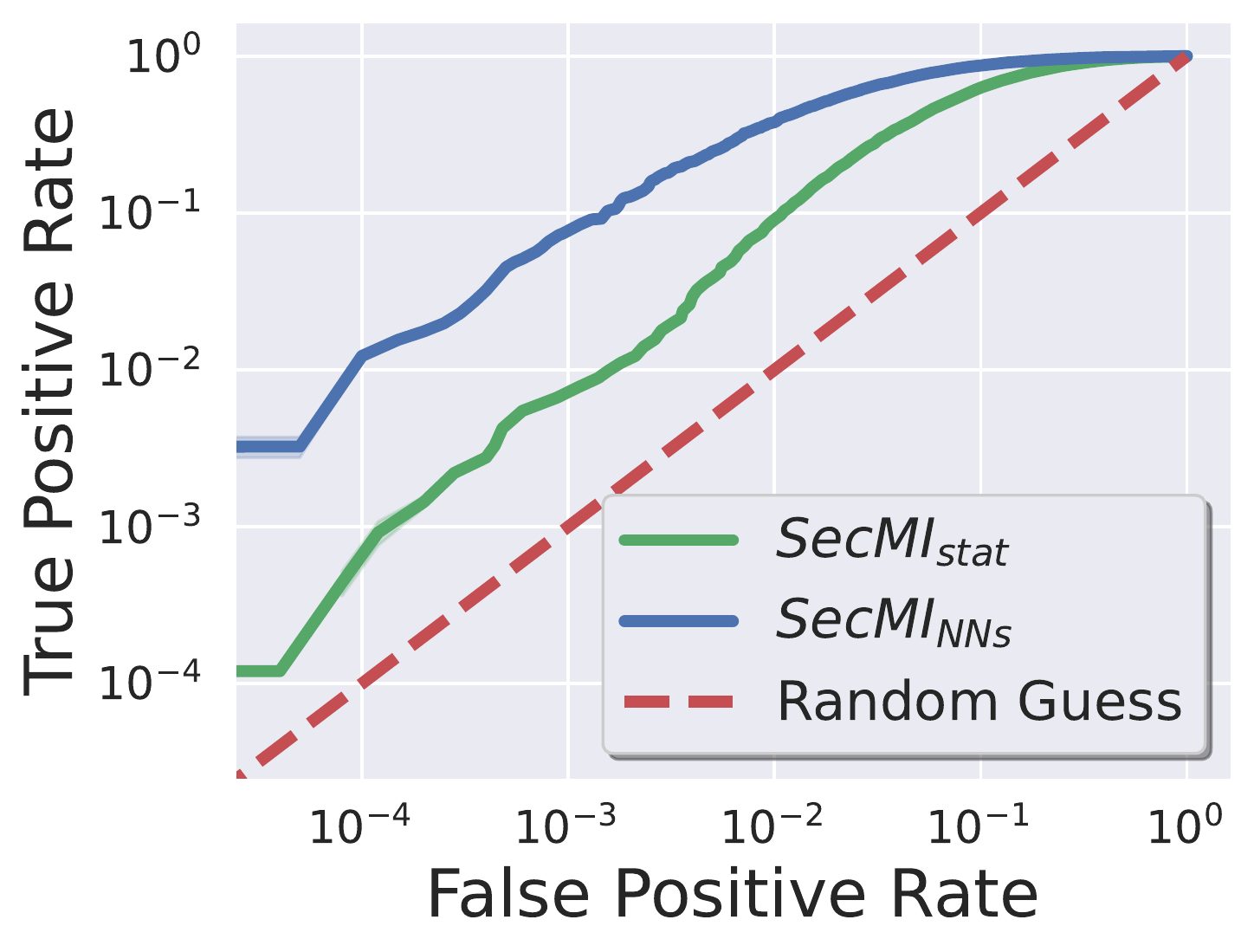}
                \caption{Log-scaled ROC on CIFAR10.}
        \end{subfigure}%
        \begin{subfigure}[b]{0.25\textwidth}
                \includegraphics[width=\linewidth, height=3.5cm]{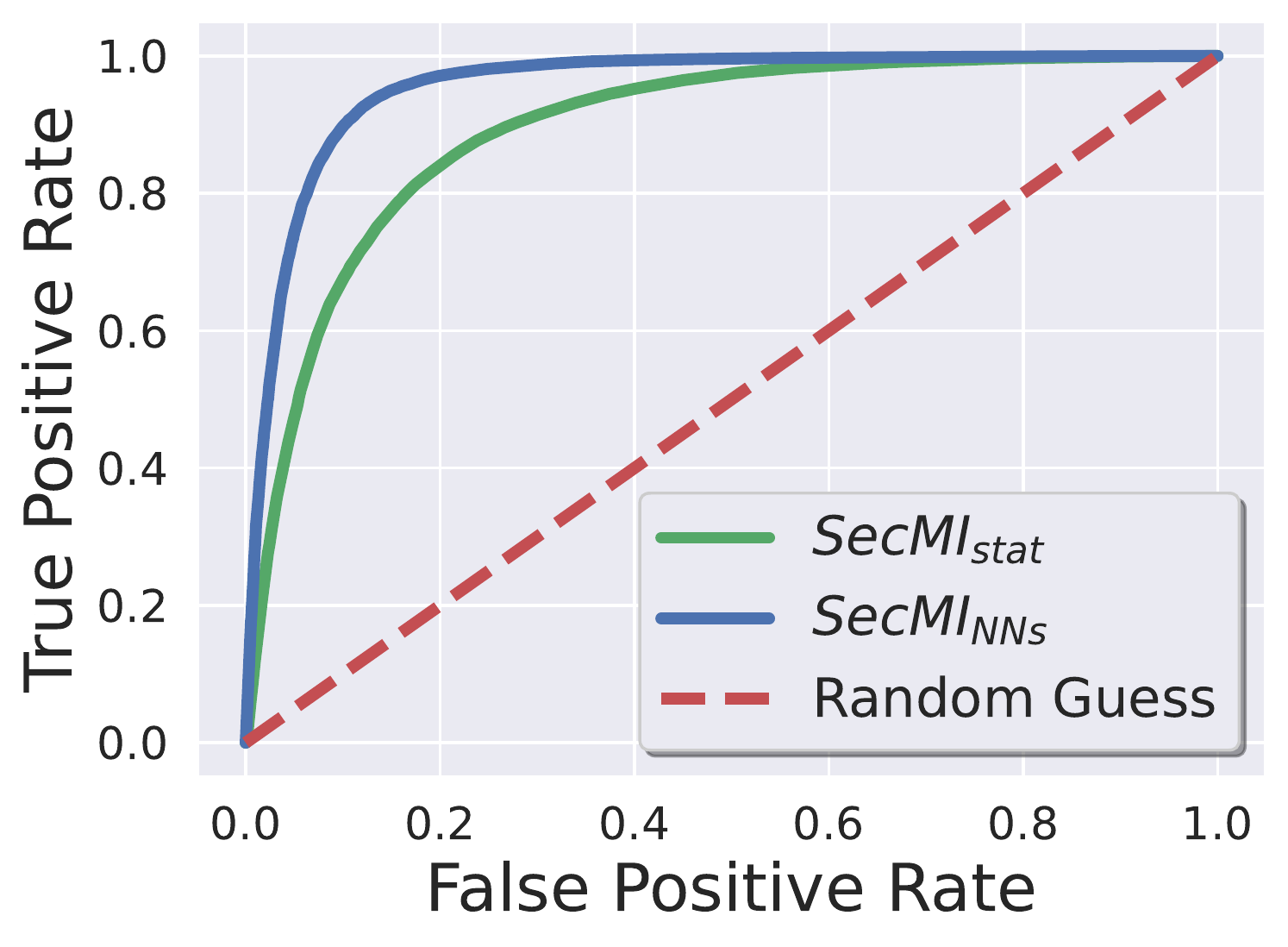}
                \caption{ROC on Tiny-IN.}
        \end{subfigure}%
        \begin{subfigure}[b]{0.25\textwidth}
                \includegraphics[width=\linewidth, height=3.5cm]{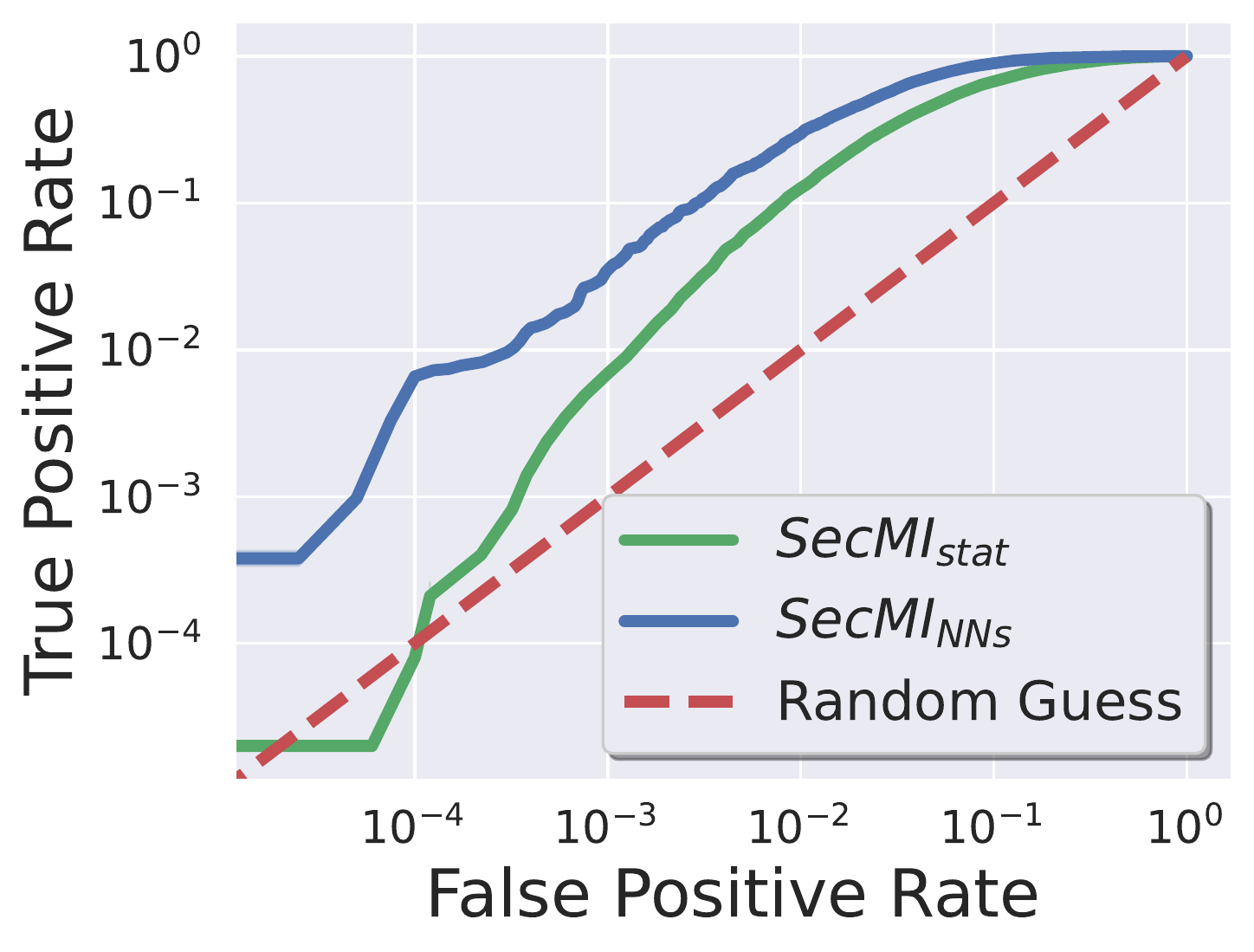}
                \caption{Log-scaled ROC on Tiny-IN.}
        \end{subfigure}
        \caption{ROC curves of SecMI on CIFAR10 and Tiny-IN datasets. The overall ROC curves show that our methods are largely effective on diffusion models. The log-scaled ROC curves indicate that our methods are capable of generating high-confidence predictions.}
        \label{fig:secmi_ROCs}
         \vspace{-2mm}
\end{figure*}

\textbf{Implementation Details.}
We train DDPM from scratch with the default hyper-parameters, except for the data splittings.
For text-to-image experiments, we adopt the HuggingFace pre-trained Stable Diffusion as the victim diffusion models. More implementation details can be found in ~\cref{sec:ldm} and ~\cref{sec:stablediffusion}.
We also adopt DDIM ($k$) to speed up the inference, where $k$ refers to the inference interval, e.g., DDIM~(10) consisting of 100 denoising steps when $T$=1,000.
We set $t_{\textsc{SEC}}$ to 100 for all the experiments.
For attack model $f_{\mathcal{A}}$, we choose ResNet-18 as the backbone and adopt 20\% of $D_{\textit{M}}$ and $D_{\textit{H}}$ as its training samples.
$f_{\mathcal{A}}$ is trained in 15 epochs with a learning rate of 0.001 and batch size of 128.

\subsection{Comparison to Baselines}
We take GAN-leaks as the baseline method. We train DDPM on four datasets, including CIFAR-10/100, STL10-U, Tiny-IN, and summarize AUCs and ASRs in~\cref{tab:ddpm_4ds_auc_asr}. It is shown that SecMI accurately infers most of the memberships for samples from these datasets, i.e., 81.0\% and 88.9\% average ASRs for SecMI$_{stat}$ and SecMI$_{\textit{NNs}}$. Compared with naive statistical inference, training a neural network as the inference strategy can significantly improve performance (SecMI$_{\textit{NNs}}$ outperforms SecMI$_{\textit{stat}}$ by over 7\%.). In~\cref{fig:secmi_dists}, we show that $t$-\textit{error} is a qualified extractor to distinguish member samples and hold-out samples.

As pointed out by~\cite{carlini2022membership}, the risk of correctly inferring membership is greater than that of being inferred incorrectly for some scenarios, e.g., medical data. Therefore, we also consider the TPR at very low FPR, e.g., 1\% FPR and 0.1\% FPR in~\cref{tab:ddpm_4ds_tpr}. It is shown that SecMI achieves notable TPR in both of the two evaluations, which proves its effectiveness. 
We present the overall ROC curves and the log-scaled ROC curves in~\cref{fig:secmi_ROCs}.

\subsection{Ablation Study}\label{sec:ablation_study}
\begin{figure}[ht]
    \centering
    \includegraphics[width=0.47\textwidth]{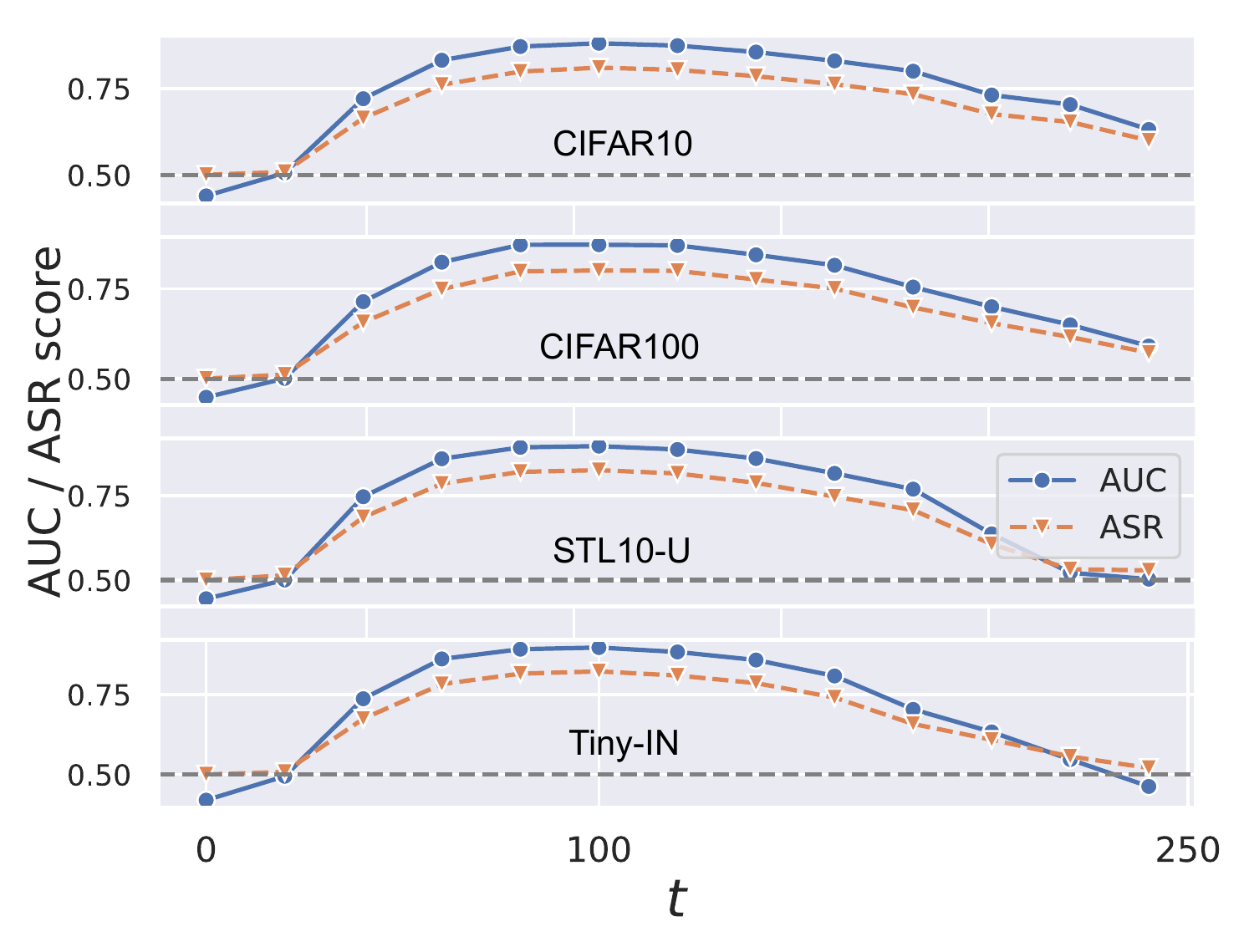}
    \caption{AUC and ASR of SecMI$_{stat}$ v.s. timestep, among four datasets. The attack performances are stable and not sensitive to the selection of timestep $t_{\textsc{SEC}}$.}
    \label{fig:auc_asr_cueves}
\end{figure}

\begin{figure}[ht]
    \centering
    \includegraphics[width=0.47\textwidth]{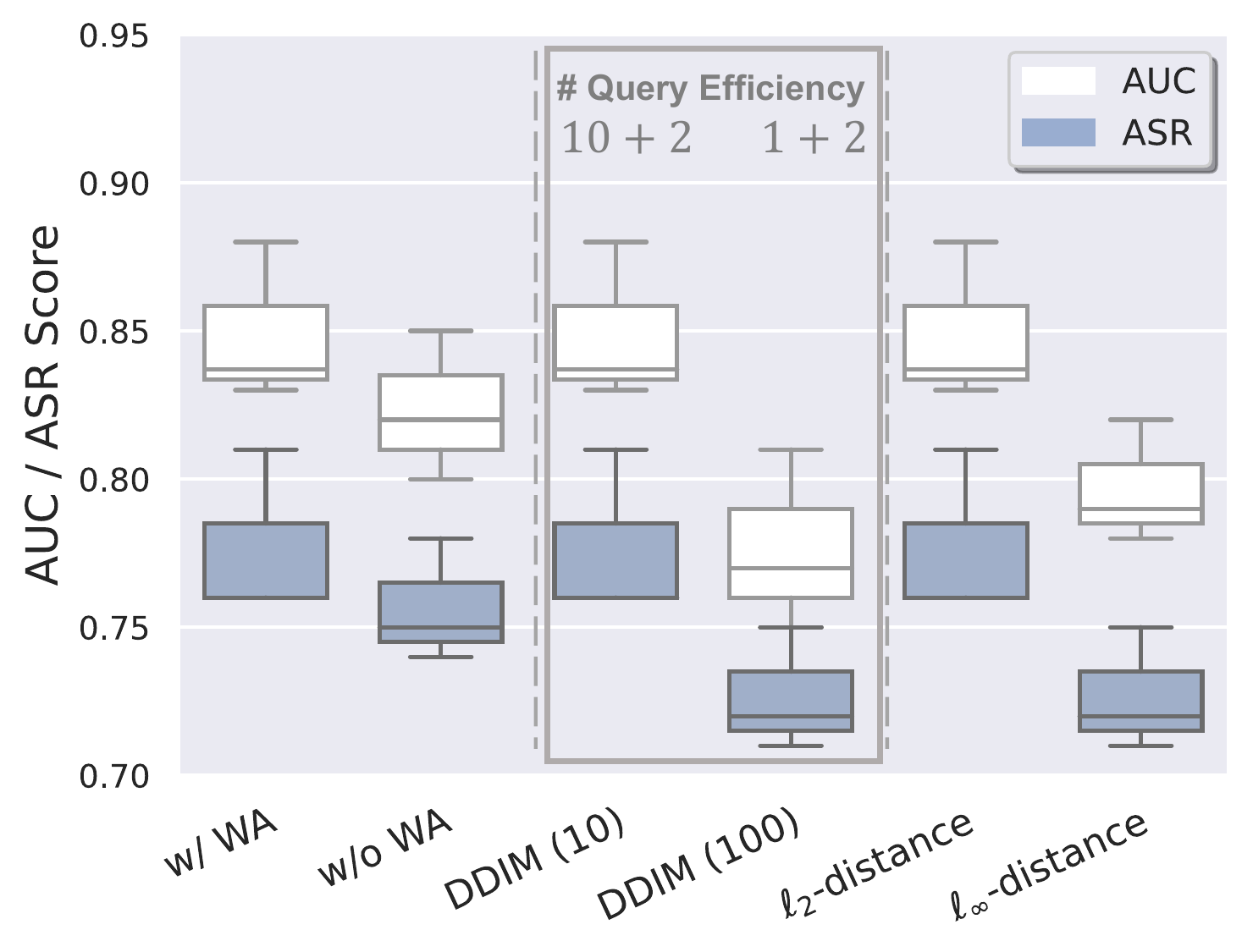}
    \caption{Ablation studies of hyper-parameters and the sensitivity to the randomness. Models are evaluated by SecMI$_{stat}$ on the CIFAR-10 dataset.}
    \label{fig:boxplot_asr_auc}
\end{figure}

We study how hyper-parameters, such as $t_{\textsc{SEC}}$, DDIM inference interval $k$, and the distance measurement $d(\cdot, \cdot)$ used in $\tilde{\ell}_{t, \vx}$, affect the performance of SecMI. Besides, since Weights Averaging (WA) is one of the most common techniques in training diffusion models, we also analyze how WA affects privacy leaking.
For each experiment, we run 5 trials with random data splittings. Results are summarized in~\cref{fig:auc_asr_cueves} and~\cref{fig:boxplot_asr_auc}.

Generally, we show that our algorithm is stable with limited variance, e.g., $\leq$ 0.05 for both AUC and ASR over all trials.
For timestep $t_{\textsc{SEC}}$, although we determine $t_{\textsc{SEC}}$ empirically, we show that the attack performances are not sensitive to specific timesteps. As long as $50 \leq t_{\textsc{SEC}} \leq 150$, the attack will be effective.
This property also generalizes well to other experiments.
$k$ affects the query efficiency and we show that SecMI still achieves remarkable adversary even with only 3 queries.  
In terms of WA, an interesting phenomenon is that WA brings a certain degree of privacy leaking, which alarms the community to carefully select the training protocols. For distance measurement, the sensitivity to the estimation error also affects the performance of our methods.

\begin{table}[ht]
    \centering
    \caption{Evaluations in resisting data augmentations.}
    \label{tab:defense}
    \vspace{-2mm}
     \resizebox{\columnwidth}{!}{
     \begin{threeparttable}
     \begin{tabular}{lcccc}
        \toprule
         & \multicolumn{2}{c}{SecMI$_{\textit{stat}}$} & \multicolumn{2}{c}{SecMI$_{\textit{NNs}}$} \\
         \cmidrule(lr){2-3}
         \cmidrule(lr){4-5}
        \multicolumn{1}{c}{Method} & ASR$\uparrow$  & AUC$\uparrow$  & ASR$\uparrow$  & AUC$\uparrow$  \\
        \midrule
         No Augmentation & 0.964 & 0.912 & 0.972 & 0.928 \\
         \midrule
         w/ RandomHorizontalFlip$^\dagger$ & 0.811 & 0.881 & 0.888 & 0.951 \\
         w/ Cutout~\cite{devries2017improved} & 0.961 & 0.908 & 0.992 & 0.973 \\
         \midrule
         \textcolor{gray}{w/ RandAugment}~\cite{cubuk2020randaugment}$^\ddagger$ & \textcolor{gray}{0.533} & \textcolor{gray}{0.530} & - & - \\ 

        \bottomrule
    \end{tabular}
    \begin{tablenotes}
        \item $^\dagger$ RandomHorizontalFlip is the default augmentation for diffusion models.
        \item $^\ddagger$ RandAugment makes diffusion models fail to converge.
    \end{tablenotes}
         
     \end{threeparttable}
    
    }
\end{table}

\subsection{SecMI v.s. Defense}
Recall that MIAs primarily benefit from overfitting~\cite{cornellMI,yeom2018privacy,salem2018mlleaks}. We investigate how data augmentation, one of the most popular methods in resisting overfitting, affects the performance of {SecMI}.
Specifically, we study Cutout~\cite{devries2017improved}, RandomHorizontalFlip. Results are summarized in~\cref{tab:defense}.
It is shown that the attack performance increased significantly when no augmentation is applied to the diffusion model. After applying mild data augmentation, e.g., Cutout and RandomHorizontalFlip, the ASR and AUC get decreased to a certain degree.

We also try to examine SecMI with stronger privacy-preserving methods and training tricks, including DP-SGD~\cite{abadi2016deep}, $\ell_2$ regularization, and stronger data augmentation such as RandAugment~\cite{cubuk2020randaugment}. We follow the same training settings as before but solely apply these techniques during model training. Experimental results show that DDPM training with these strong defense methods even failed to converge. For instance, with $\ell_2$ regularization, the trained DDPM can only generate random and meaningless information. Similarly, applying RandAugment will make the generated data full of distortions, which makes it valueless to evaluate our methods. Some generated images under the above settings are provided in~\cref{appendix:failed_defense}.

\subsection{Evaluations on Latent Diffusion Models (LDMs)}\label{sec:ldm}
We conduct experiments on text-to-image diffusion models in this section. Specifically, we adopt the Huggingface pre-trained stable-diffusion-v1-4 as the backbone and fine-tune 15,000 and 150,000 steps over Pokemon and COCO2017-Val, respectively. Both two datasets are not considered during the Stable Diffusion pre-training phase. We study the sensitivity to ground-truth prompts, i.e., whether adopting empty prompts or prompts generated from other sources (e.g., BLIP~\cite{li2022blip}) during the attack will affect performances.

Results are summarized in~\cref{tab:ldm}. It is shown that our method achieves superior attack performances on both two datasets, which proves that {SecMI} generalizes well to the sophisticated diffusion models. In terms of the sensitivity to ground-truth prompts, we show that the actual influences vary among different tasks. For the Pokemon dataset, only a marginal drop (around 0.03) can be observed when equipped with empty prompts. We believe this is because the Pokrmon dataset has highly centralized images and text (they are all the characters of PokemonGo with a very uniform style), which makes the generation less sensitive to the prompt.

\begin{figure}[t]
    \centering
    \includegraphics[width=0.47\textwidth]{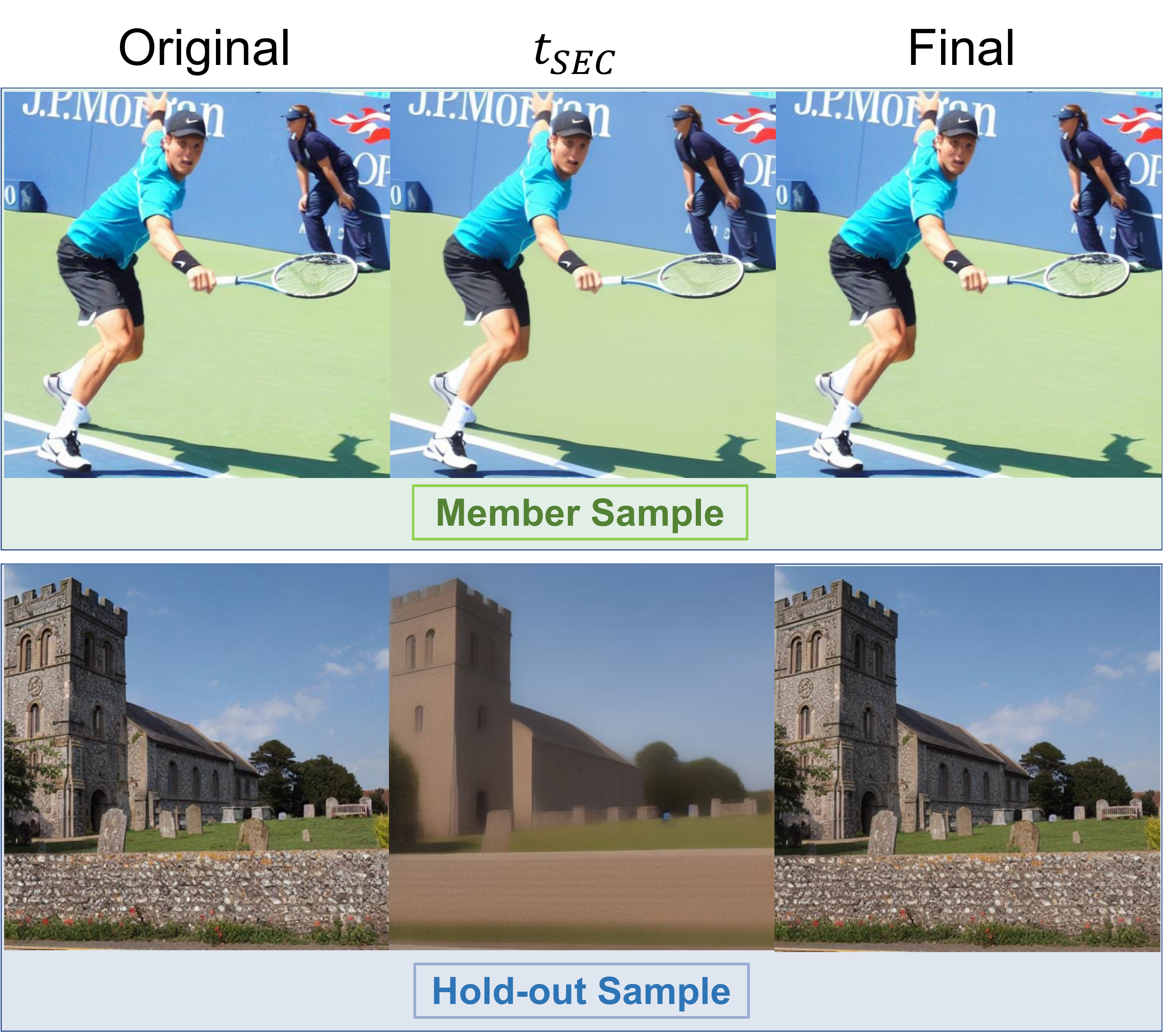}
    \caption{Visualization results of image reconstruction from a member sample and a hold-out sample at the $t_{\textsc{SEC}}$-th step and the final step.}
    \label{fig:visualization}
\end{figure}

In contrast, we show that empty prompts will largely affect (around 0.15) the attack performance over the COCO2017-val set. 
However, with pseudo-prompts generated by BLIP, this drop can be almost mitigated (around 0.05). We believe it is because of the diversity and variety of the COCO2017-val images (i.e., diverse backgrounds, objects, and events), which makes the generation more sensitive to the prompt.
This indicates that our SecMI is applicable when the attacker can’t get access to the ground-truth prompts.

We also provide visualization results in~\cref{fig:visualization}, by directly reconstructing the reverse results at the $t_{\textsc{SEC}}$-th step and the final step, to show the qualitative differences between member samples and hold-out samples.

\subsection{Evaluations on Stable Diffusion}\label{sec:stablediffusion}
We conduct experiments on the original Stable Diffusion, i.e., stable-diffusion-v1-4 and v1-5 provided by Huggingface, without further fine-tuning or other modifications. Specifically, we randomly sample 2500 images from Laion-aesthetic images with aesthetic scores greater than 5 (Laion-aesthetic-5plus) as the member set and randomly sample 2500 images from COCO2017-val as the hold-out set.

As shown in~\cref{tab:sd}, our method achieves notable attack performances on both stable-diffusion-v1-4 and v1-5. It indicates that SecMI is still effective for large-scale pre-training scenarios, which sheds light on applying our method in the real world.

\section{Conclusion}
In this paper, we investigate whether diffusion models are vulnerable to MIAs. Specifically, we first summarize existing MIAs and show that most of them are largely ineffective for diffusion models. Then, to mitigate this gap, we propose Step-wise Error Comparing Membership Inference (SecMI), based on the step-wise posterior matching of diffusion models. We evaluate SecMI on both standard diffusion models, e.g., DDPM, and state-of-the-art text-to-image models, e.g., Stable Diffusion. Experimental results across multiple datasets demonstrate the effectiveness of our method. 

Our research reveals that current diffusion models suffer from serious privacy issues. We hope that our work can inspire the community and encourage more privacy considerations.

\begin{table}[t]
\caption{Evaluations on LDMs. Models are evaluated by SecMI$_{stat}$.}
    \label{tab:ldm}
    \vspace{-2mm}
    \centering
    \adjustbox{width=1\columnwidth}{
    \resizebox{\columnwidth}{!}{
    \begin{tabular}{ccccc}
        \toprule
        Dataset & Prompt & ASR$\uparrow$  & AUC$\uparrow$  & TPR@1\%FPR (\%) $\uparrow$   \\
        \midrule
          \multirow{2}{*}{Pokemon} & Ground-truth & 0.821 & 0.891 & 7.20 \\
           & Empty & 0.782	& 0.860	& 11.06 \\
        \midrule
          \multirow{3}{*}{COCO2017-val} & Ground-truth & 0.803 & 0.875 & 13.98 \\
          & Empty & 0.663 & 0.720 & 6.04 \\
          & BLIP~\cite{li2022blip} & 0.750 & 0.820 & 9.40 \\
          
         \bottomrule
    \end{tabular}
    }
    }
    
\end{table}

\begin{table}[t]
\caption{Evaluations on Stable Diffusion. Models are evaluated by SecMI$_{stat}$.}
\label{tab:sd}
    \centering
    \resizebox{\columnwidth}{!}{
    \begin{threeparttable}
        \begin{tabular}{cccc}
    \toprule
    Victim Model & AUC & ASR & TPR@1\%FPR (\%)\\ 
    \midrule
    stable-diffusion-v1-4$^\dagger$ & 0.707 & 0.664 & 18.47 \\
    stable-diffusion-v1-5 & 0.701 & 0.661 & 18.58\\
    \bottomrule
    \end{tabular}
    \begin{tablenotes}
        \item $^\dagger$ stable-diffusion-v1-4 is trained in less than 1 epoch over the Laion-aesthetic-5plus dataset. The real attack performance should be scaled.
    \end{tablenotes}
    \end{threeparttable}
    }

\end{table}

\section*{Limitations}
Despite SecMI showing great attack performances on various diffusion models and datasets, there are some limitations when applied in the physical world: 1) diffusion models are normally provided as black-box API services and it is less possible the attacker will access the intermediate results of the victim diffusion models, while SecMI requires the access to these results; 2) SecMI is only evaluated on public data, the effectiveness and sensitivity to a given demographic (or a subgroup of a dataset) is not investigated yet; 3) although MIA is one of the most common privacy concern in academia, the scope of MIA is limited in the real world.

\section*{Ethics and Broader Impacts}
This paper proposes a membership inference attack algorithm, which is a threat to privacy for current diffusion models. To mitigate any possible abuse caused by this paper, all the experiments are conducted on public datasets and common model architectures. Figures presented in this paper are licensed under the Creative Commons 4.0 License, which is allowed to be distributed.

Despite there are limitations when applying our method in the physical world, it is still possible that our paper will cause privacy risks. However, we believe our paper acts more like an alert to the generative model community. We encourage more privacy and security considerations before releasing diffusion models to the public.


\bibliography{main}
\bibliographystyle{icml2023}

\newpage
\appendix
\onecolumn
\appendix



\section{Generalization to Latent Diffusion Model (LDM)}\label{appendix:generalize_various_diffusion}
For the Latent Diffusion Model (LDM), the calculation of $t$-\textit{error} is similar to DDPM, except the intermediate latent variables are in the latent space and the reverse process is conditioned by text embeddings.
Specifically, we denote by $V$ the Variational Autoencoders (VAEs) utilized to encode the original image into the latent space, i.e., $\vv_0 = V(\vx_0), \vx_0 \sim D$, and denote by $C$ the text condition.
The diffusion process and the denoising process can be derived as:
\begin{equation}
\begin{aligned}
q(\vv_{t}|\vv_{t-1}) = \mathcal{N}(\vv_{t}; \sqrt{1 - \beta_t}\vv_{t-1},  \beta_t \textbf{I}) \,\,\, \,\,\,\\
    p_{\vtheta}(\vv_{t-1}|\vv_{t}) = \mathcal{N}(\vv_{t-1};\mu_{\theta}(\vv_t, t, C),  \Sigma_{\theta}(\vv_t, t)).
\end{aligned}
\end{equation}
Then $t$-\textit{error} can be rewrite as:
\begin{equation}
    \tilde{\ell}_{t, \vv_0} = || \psi_{\vtheta} ( \phi_{\vtheta}(\tilde{\vv}_t, t, C), t, C) - \tilde{\vv}_{t} ||^2,
\end{equation}
where we reuse the symbols $\phi_{\vtheta}$ and $\psi_{\vtheta}$ as the deterministic reverse and sampling regarding $\vv_t$:
\begin{equation}
\begin{aligned}
    \vv_{t+1} & = \phi_{\vtheta}(\vv_t, t, C) \\
    &= \sqrt{\bar{\alpha}_{t+1}}f_\vtheta(\vv_t, t, C) + \sqrt{1 - \bar{\alpha}_{t+1}}\epsilon_{\vtheta}(\vv_t, t, C),
\end{aligned}
\end{equation}
\begin{equation}
\begin{aligned}
    \vv_{t-1} & = \psi_{\vtheta}(\vv_t, t, C) \\
    & = \sqrt{\bar{\alpha}_{t-1}}f_\vtheta(\vv_t, t, C) + \sqrt{1 - \bar{\alpha}_{t-1}}\epsilon_{\vtheta}(\vv_t, t, C),
\end{aligned}
\end{equation}
where
\begin{equation}
    f_\vtheta(\vv_t, t, C) = \frac{\vv_t - \sqrt{1 - \bar{\alpha}_t}\epsilon_\vtheta(\vv_t, t, C)}{\sqrt{\bar{\alpha}_t}}.
\end{equation}   

\section{Adopted Diffusion Models and Datasets}\label{appendix:adopted_model_datasets}
Adopted diffusion models and datasets, along with the data splittings, are summarized in~\cref{tab:dataset_diffusion_maps}.

\begin{table*}[ht]
    \caption{Adopted diffusion models and datasets.}
    \label{tab:dataset_diffusion_maps}
    \centering
    \resizebox{0.8\columnwidth}{!}{
    \begin{threeparttable}
    \begin{tabular}{cccccc}
        \toprule
        Model & Dataset & Resolution & \# Member & \# Hold-out & Cond.\\
        \midrule
        \multirow{4}{*}{DDPM} & CIFAR-10 & 32 & 25,000 & 25,000 & - \\
        & CIFAR-100 & 32 & 25,000 & 25,000 & -\\ 
        & STL10-U & 32 & 50,000 & 50,000 & -\\ 
        & Tiny-ImageNet & 32 & 50,000 & 50,000 & -\\ 
        \midrule
        \multirow{2}{*}{Latent Diffusion Model} & Pokemon & 512 & 416 & 417 & text\\
        & COCO2017-Val & 512 & 2,500 & 2,500 & text\\
        \midrule
        Stable Diffusion V1-4/5 & Laion & 512 & 2500 & 2500 & text \\
        \bottomrule
    \end{tabular}
    \end{threeparttable}
    }
\end{table*}

\section{Failed Defensive Training Results}\label{appendix:failed_defense}

\begin{figure}
    \centering
    \includegraphics{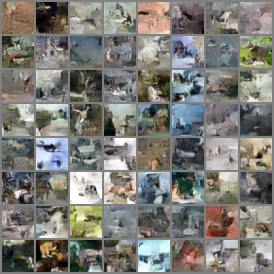}
    \caption{Sampling results at the 800,000th step from DDPM trained with $\ell_2$ regularization.}
    \label{fig:l2}
\end{figure}

\begin{figure}
    \centering
    \includegraphics{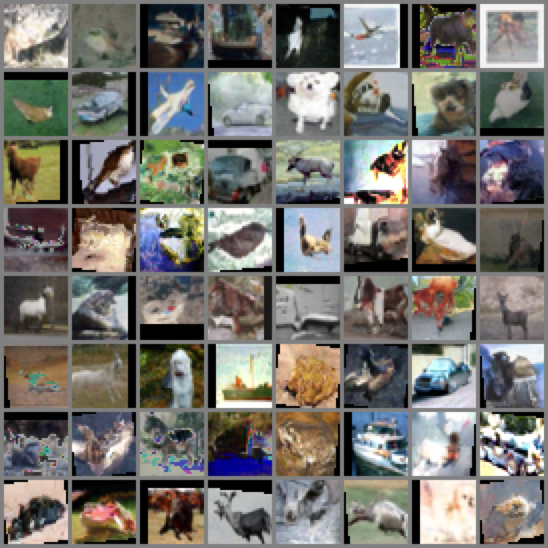}
    \caption{Sampling results at the 800,000th step from DDPM trained with RandAugment.}
    \label{fig:random_arg}
\end{figure}

We show some sampling images of defensive training results from Figure~\ref{fig:l2} and~\ref{fig:random_arg}. The generated results are vague and unrealistic due to the aggressive regularization or data argumentation applied to the training process. Subsequently, images from whatever member or hold-out set will be reconstructed at a low quality, rendering them unworthy to conduct MIA.

\end{document}